\documentclass[10pt]{article}
\usepackage[preprint]{tmlr}

\usepackage{amsmath,amssymb,amsfonts,amsthm}
\usepackage{algorithm}
\usepackage{algpseudocode}
\usepackage{graphicx}
\usepackage{booktabs}
\usepackage{array}
\usepackage{multirow}
\usepackage{xcolor}
\usepackage{colortbl}
\definecolor{fsshade}{gray}{0.92}
\usepackage{url}
\usepackage[hidelinks]{hyperref}

\theoremstyle{plain}
\newtheorem{theorem}{Theorem}
\newtheorem{lemma}{Lemma}
\newtheorem{corollary}{Corollary}
\newtheorem{proposition}{Proposition}
\theoremstyle{remark}

\newcommand{\bthe}{\boldsymbol{\theta}}
\newcommand{\E}{\mathbb{E}}
\newcommand{\Fil}{\mathcal{F}}
\newcommand{\Emax}{E_{\max}}
\newcommand{\Lc}{\mathcal{L}}
\newcommand{\fedstop}{\textsc{FedStop}}

\begin{document}

\title{Local Epochs as Optimal Stopping:\\ Theory and Evidence in Federated Learning}

\author{\name Ke Xiao \email ke.xiao@glasgow.ac.uk \\
      \addr School of Computing Science, University of Glasgow
      \AND
      \name Qiyuan Wang \email qiyuan.wang@glasgow.ac.uk \\
      \addr School of Computing Science, University of Glasgow
      \AND
      \name Christos Anagnostopoulos \email christos.anagnostopoulos@glasgow.ac.uk \\
      \addr School of Computing Science, University of Glasgow}

\maketitle
\suppressfloats[t]

\begin{abstract}
How long should each client train before communicating? Federated learning usually answers with a fixed, hand-tuned number of local epochs, yet an extra epoch helps only while the loss it removes outweighs the client drift it adds. We cast local training as finite-horizon optimal stopping: a client trains while the next epoch's expected loss reduction exceeds its priced squared drift. We prove that this one-stage look-ahead threshold is optimal whenever stopping decisions are nested, derive closed forms for geometric and power-law returns, and show that an estimated rule loses utility only through wrong-signed decisions. Epoch-level evidence supports the model's structure (geometric diminishing returns and additive drift) and reveals a simple regularity: the stage of training, not the client, sets the budget. The resulting rule, \fedstop{}, is never more than one point below tuned FedAvg in mean accuracy. Where its budgets vary, it raises worst-decile accuracy by 4--8 points over tuned FedAvg and an adaptive-interval baseline, and every one of these gains remains significant after a Holm correction. Against the client-side stopping rules ALT and FLASH, it has the highest CIFAR-10 worst decile on every seed and is level elsewhere. It stays stable where per-client early stopping collapses, and each stopping decision takes constant time.
\end{abstract}

\section{Introduction}

In federated learning (FL) with federated averaging (FedAvg), each round the server broadcasts the global model, a sample of clients train it locally for $E$ epochs, and the server averages the returned models \citep{mcmahan2017}. The local epoch count $E$ governs a basic trade-off. More local work reduces how often clients must communicate. Under statistical heterogeneity, however, it also lets each local model drift toward its own optimum and away from the global one, which degrades the aggregate \citep{li2020fedprox,karimireddy2020}.

In practice $E$ is a tuned constant, and a fragile one. In a grid that also varies the learning rate, the best fixed configuration is two epochs on CIFAR-10 but one on CIFAR-100, where two epochs diverge (Section~\ref{sec:consequences}). More fundamentally, a tuned constant says nothing about \emph{why} a budget works, how it should change as training proceeds, or whether it should differ across clients.

We take a modelling view. Every local epoch is an investment: after it, a client observes what it bought, a reduction in its local loss, and what it paid, a further displacement from the round's starting model. Returns diminish as the local loss flattens while the displacement keeps accruing, so deciding when to stop can be cast as a finite-horizon optimal stopping problem. A model of this kind is worth having only to the extent that it describes real local training, so we derive what it predicts and test those predictions against epoch-level logs. To our knowledge, this is the first formulation of a client's choice of local epochs as an optimal stopping problem. The resulting rule, \fedstop{} (\textbf{Fed}erated optimal \textbf{Stop}ping), runs locally on each client and ends its training as soon as another epoch is no longer worth its drift.

\textbf{Contributions.}
\begin{enumerate}
\item \emph{Model.} Local-epoch selection as drift-priced optimal stopping, with a utility motivated by a one-round aggregation bound.
\item \emph{Theory.} We prove that a one-stage threshold rule is optimal when stopping decisions are nested, including correlated gains within a margin. We derive closed forms with exact one-epoch conditions for geometric and power-law returns, also under declining drift. A regret identity shows that a rule using estimates loses utility only through wrong-signed decisions.
\item \emph{Evidence.} We derive five observable consequences of the model, such as diminishing returns per epoch and drift that adds up across epochs, and check each against epoch-level training logs. We report where the model holds and where it does not, and find that budgets follow the stage of training rather than the client.
\item \emph{Practice.} We compare \fedstop{} with the two client-side stopping rules closest to it, ALT \citep{shenaj2025alt} and FLASH \citep{panchal2023flash}, and with FedAvg \citep{mcmahan2017}, FedNova \citep{wang2020fednova}, the adaptive interval of \citet{wang2019adaptive}, FedProx \citep{li2020fedprox} and per-client early stopping. Random-swap and equal-split controls show where its gains come from. \fedstop{} costs at most one point of mean accuracy against tuned FedAvg. Under strong label skew, it lifts the worst decile of clients by up to eight points over tuned FedAvg, a tail that fixed budgets match only with at least twice as many local epochs, if at all. Among the client-side rules, it has the highest CIFAR-10 worst decile, ahead of ALT and FLASH on every seed, and it is level with them on the naturally partitioned data. It remains stable where per-client early stopping fails. The decision itself is lightweight: constant time per epoch and a few hundred scalars of memory per client, independent of model size.
\end{enumerate}

\section{Related Work}
\label{sec:related}

\textbf{Client drift.} FedProx adds a proximal penalty to each local objective \citep{li2020fedprox}, SCAFFOLD corrects local directions with control variates \citep{karimireddy2020}, and FedNova normalises updates by the number of local steps to remove objective inconsistency \citep{wang2020fednova}. These methods change the local objective or the aggregation rule and are complementary to choosing the training duration. \fedstop{} uses FedNova-style normalisation directly.

\textbf{Local computation.} The effect of the number of local steps on convergence is well studied for local SGD \citep{stich2019,khaled2020,woodworth2020}, including step counts that vary across rounds \citep{qin2023} and the dependence of the surrogate objective on local work \citep{charles2021surrogate,patel2024limits}. Wang et al.\ adapt a shared aggregation interval under a resource budget \citep{wang2019adaptive}. These analyses bound or tune one budget per round. We instead model the budget as a client-side stopping decision and test the model's assumptions directly.

\textbf{Client-side and global stopping.} Two methods also set each client's number of epochs, by different signals and for different ends. ALT stops a client once its representations diverge from the global model's \citep{shenaj2025alt}. FLASH addresses concept drift: it pairs a drift-aware server optimiser with a client rule that stops once validation loss stops falling \citep{panchal2023flash}. Neither models the trade-off it decides. \fedstop{} derives its threshold from an explicit stopping model with a priced drift cost, and Section~\ref{sec:consequences-D} compares the three rules on CIFAR-10, FEMNIST and Shakespeare. At the server, stopping decisions take two forms: when to close a round given the client updates received so far, posed as optimal stopping \citep{jiang2020optimal}, and when to terminate training altogether \citep{mahmoudi2022fedcau,niu2024flrce}. \fedstop{} moves the stopping decision to the client, where it sets how long each client trains within a round. We report worst-client accuracy as an outcome, but unlike fairness-oriented methods \citep{li2020qffl,mohri2019afl} we do not optimise for it.

\section{Preliminaries}
\label{sec:prelim}

\subsection{Federated learning}\label{sec:prelim-A}

Client $i$ holds $n_i$ examples and a local objective $F_i$. The population objective is $F=\sum_ip_iF_i$ with $p_i=n_i/\sum_jn_j$. In round $t$ the server broadcasts $\bthe^{(t)}$ to a sampled cohort $\mathcal S_t$ with aggregation weights $q_i=n_i/\sum_{j\in\mathcal S_t}n_j$. Each client starts from $\bthe_i(0)=\bthe^{(t)}$ and runs local epochs of $S_i=\lceil n_i/B\rceil$ minibatch SGD steps at batch size $B$. In FedAvg every client runs $E$ epochs and the server forms $\bthe^{(t+1)}=\sum_{i\in\mathcal S_t}q_i\bthe_i(E)$. When clients execute different numbers of steps $K_i$, plain averaging implicitly upweights those that train longer and optimises an inconsistent objective. Step normalisation \citep{wang2020fednova} corrects this with
\begin{equation}
\bthe^{(t+1)}=\bthe^{(t)}+\bar K\sum_{i\in\mathcal S_t}q_i\,\frac{\bthe_i-\bthe^{(t)}}{K_i},\qquad \bar K=\sum_{i\in\mathcal S_t}q_iK_i ,
\label{eq:fednova}
\end{equation}
which reduces to FedAvg when all $K_i$ are equal. The distance $\|\bthe_i-\bthe^{(t)}\|$ that a client travels from the broadcast is its \emph{client drift} \citep{karimireddy2020}.

\subsection{Finite-horizon optimal stopping}\label{sec:prelim-B}

We recall the elements of optimal stopping used in this paper \citep{chow1971,peskir2006}. Let $(U_k)_{k=1}^{H}$ be a reward process adapted to a filtration $(\Fil_k)$, with $\E|U_k|<\infty$. A \emph{stopping time} $\tau$ takes values in $\{1,\dots,H\}$ and satisfies $\{\tau=k\}\in\Fil_k$, so the decision to stop at $k$ uses only the information available at $k$. The finite-horizon problem is to find $\sup_\tau\E[U_\tau]$.

\textbf{Backward induction.} The value of the problem is obtained by the recursion
\begin{equation}
V_H=U_H,\quad C_k=\E[V_{k+1}\mid\Fil_k],\quad V_k=\max\{U_k,C_k\},
\label{eq:snell}
\end{equation}
for $k<H$, where $C_k$ is the value of continuing optimally and $V_k$ is the Snell envelope of the reward process.

\begin{proposition}[Existence of an optimal rule \citep{chow1971}]
\label{prop:existence}
The stopping time $\tau^\circ=\min\{k:U_k\ge C_k\}$, with $C_H:=-\infty$, is optimal, and $\E[U_{\tau^\circ}]=\E[V_1]=\sup_\tau\E[U_\tau]$.
\end{proposition}

Proposition~\ref{prop:existence} guarantees that our stopping problem (Section~\ref{sec:setup}) always has an optimal solution, but computing $C_k$ in general requires a model of the entire future. The \emph{monotone case} identifies when a much simpler rule suffices. The one-stage look-ahead rule $\tau_1=\min\{k<H:U_k\ge\E[U_{k+1}\mid\Fil_k]\}\wedge H$ compares stopping now with stopping one step later. The problem is \emph{monotone} if the sets $A_k=\{U_k\ge\E[U_{k+1}\mid\Fil_k]\}$ satisfy $A_1\subseteq A_2\subseteq\cdots\subseteq A_H$ almost surely, so that once the one-stage comparison favours stopping it continues to do so.

\begin{lemma}[Monotone case \citep{chow1971}]
\label{lem:monotone}
In a monotone finite-horizon problem, $\tau_1$ is optimal.
\end{lemma}

\section{Problem Formulation}
\label{sec:setup}

\subsection{Local progress and drift}\label{sec:setup-A}

Within round $t$, client $i$ runs local epochs $k=1,2,\dots$ from $\bthe_i(0)=\bthe^{(t)}$ (Section~\ref{sec:prelim-A}). Writing $\Lc_i(k)=F_i(\bthe_i(k))$ for the local loss, we define the marginal loss decrease, the drift, and the marginal squared drift:
\begin{align}
\Delta\Lc_i(k) &:= \Lc_i(k-1)-\Lc_i(k),\\
D_i(k) &:= \|\bthe_i(k)-\bthe^{(t)}\|,\\
\delta_i(k) &:= D_i(k)^2-D_i(k-1)^2 .
\end{align}
With $x_{k}=\bthe_i(k)-\bthe^{(t)}$ and $v_{k+1}=\bthe_i(k+1)-\bthe_i(k)$, the marginal squared drift satisfies the exact identity
\begin{equation}
\delta_i(k+1)=\|v_{k+1}\|^2+2\langle x_k,v_{k+1}\rangle ,
\label{eq:signed}
\end{equation}
so the squared drift is the sum of its per-epoch increments, $D_i(k)^2=\sum_{j\le k}\delta_i(j)$. The filtration $\Fil_k$ contains the broadcast and all local observations through epoch $k$. A client executes an epoch before observing its outcome. It cannot preview an epoch and decline its cost. We consider synchronous rounds, in which the server aggregates once every sampled client has finished. A client's budget is set by its own drift-priced utility and does not account for the time it takes to report back. A longer budget may delay the round, but delay does not enter the decision. Deadline-aware or asynchronous stopping, in which communication delay enters the utility, is beyond our scope.

\subsection{Drift-priced utility and the stopping problem}\label{sec:setup-B}

For a drift price $\lambda>0$, the net value of stopping after epoch $k$ is
\begin{equation}
U_i(k):=\Lc_i(0)-\Lc_i(k)-\lambda D_i(k)^2=\sum_{j\le k}g_i(j),
\label{eq:utility}
\end{equation}
with net increments $g_i(j)=\Delta\Lc_i(j)-\lambda\delta_i(j)$. Each client solves
\begin{equation}
\sup_{\tau\in\mathcal T}\ \E\big[U_i(\tau)\big],
\label{eq:stopping}
\end{equation}
where $\mathcal T$ is the set of $(\Fil_k)$-stopping times with values in $\{1,\dots,\Emax\}$. The first epoch is compulsory, so every participating client returns a trained model.

\subsection{Motivation from an aggregation bound}\label{sec:setup-C}

The utility \eqref{eq:utility} is not an ad hoc penalty: it is motivated by a lower bound on the descent of the aggregate. Let $\Delta_i=\bthe_i(\tau_i)-\bthe$ be client $i$'s update from broadcast $\bthe$, $P_i=F_i(\bthe)-F_i(\bthe+\Delta_i)$ its local progress, and $D_i=\|\Delta_i\|$.

\begin{lemma}[One-round aggregation bound]
\label{lem:descent}
Suppose the target $F$ has an $L$-smooth upper quadratic support along the aggregate segment, each client satisfies $F_i(\bthe+\Delta_i)\ge F_i(\bthe)+\langle\nabla F_i(\bthe),\Delta_i\rangle+\tfrac{\nu}{2}\|\Delta_i\|^2$ for some $\nu\le L$ (possibly negative), and $\|\nabla F(\bthe)-\nabla F_i(\bthe)\|\le\zeta$. Then the plain aggregate $a=\sum_iq_i\Delta_i$ satisfies, for any data-dependent stopping times,
\begin{equation}
F(\bthe)-F(\bthe+a)\ \ge\ \sum_iq_i\Big(P_i-\tfrac{L-\nu}{2}D_i^2-\zeta D_i\Big).
\label{eq:descent}
\end{equation}
\end{lemma}

\begin{proof}
By the upper support, $F(\bthe+a)-F(\bthe)\le\langle\nabla F(\bthe),a\rangle+\frac L2\|a\|^2$. Writing $\langle\nabla F,a\rangle=\sum_iq_i\langle\nabla F_i,\Delta_i\rangle+\sum_iq_i\langle\nabla F-\nabla F_i,\Delta_i\rangle$, the lower support gives $\langle\nabla F_i,\Delta_i\rangle\le-P_i-\frac\nu2D_i^2$ and Cauchy--Schwarz bounds the second sum by $\zeta\sum_iq_iD_i$. Jensen's inequality gives $\|a\|^2\le\sum_iq_iD_i^2$. Combining the three bounds yields \eqref{eq:descent}. Every step holds pathwise at the realised endpoints.
\end{proof}

With $\lambda=(L-\nu)/2$, the quadratic part of each client's term in \eqref{eq:descent} is exactly $U_i(\tau_i)$. A client that maximises its drift-priced utility therefore maximises the quadratic part of its contribution to a guaranteed decrease of the aggregate loss. The remaining term $\zeta D_i$ penalises drift further. Because $L$ and $\nu$ are not available for neural networks, and because step-normalised aggregation rescales both terms, we treat $\lambda$ as a hyperparameter selected on validation data.

\section{FedStop: Optimal Stopping for Local Training}
\label{sec:rule}

\subsection{Threshold optimality}\label{sec:rule-A}

Fix a client and round and drop the index $i$. Let $m_j:=\E[g(j)\mid\Fil_{j-1}]$ be the conditional expected net gain of epoch $j$.

\begin{theorem}[Threshold optimality]
\label{thm:threshold}
Suppose $U(k)$ is integrable and each $m_j$ is deterministic, with $m_2,\dots,m_{\Emax}$ non-increasing. Then an earliest optimal stopping time for \eqref{eq:stopping} is
\begin{equation}
\tau^\star=\min\big(\{k<\Emax:\ m_{k+1}\le0\}\cup\{\Emax\}\big).
\label{eq:threshold}
\end{equation}
\end{theorem}

\begin{proof}
Since $U(k+1)=U(k)+g(k+1)$, we have $U(k)-\E[U(k+1)\mid\Fil_k]=-m_{k+1}$, so the one-stage sets of Section~\ref{sec:prelim} are $A_k=\{m_{k+1}\le0\}$ for $k<\Emax$ and $A_{\Emax}=\Omega$. They are deterministic and, because $m$ is non-increasing, nested. The problem is therefore monotone, and Lemma~\ref{lem:monotone} shows that the one-stage rule, which is \eqref{eq:threshold}, is optimal. It is the earliest optimum because ties $m_{k+1}=0$ are resolved by stopping.
\end{proof}

The rule is the one-stage look-ahead policy: train one more epoch exactly while its expected benefit exceeds its expected drift cost. In deployment the gains are estimated, and Theorem~\ref{thm:regret} below quantifies what this costs.

\subsection{Closed forms for two progress profiles}\label{sec:rule-B}

Diminishing returns are captured by a progress profile $\psi$ with $\psi(1)=1$: $\E[\Delta\Lc(j)\mid\Fil_{j-1}]=\bar\ell\,\psi(j)$ and $\E[\delta(j)\mid\Fil_{j-1}]=\bar\delta$, with $\bar\ell,\bar\delta>0$. We study two profiles. The \emph{geometric} profile, $\psi(j)=\rho^{\,j-1}$ with progress ratio $0<\rho<1$, is the decay implied by linear convergence under the Polyak--{\L}ojasiewicz condition \citep{karimi2016pl,liu2022pl}. The \emph{power-law} profile, $\psi(j)=j^{-a}$ with decay exponent $a>0$, matches the sublinear rates of stochastic gradient methods \citep{bottou2018optimization} and the power laws observed for loss against training steps \citep{kaplan2020scaling}. Both are modelling assumptions, tested in Section~\ref{sec:model-A}. We write
\begin{equation}
\kappa:=\lambda\bar\delta/\bar\ell
\label{eq:kappa}
\end{equation}
for the dimensionless exchange ratio between drift cost and first-epoch progress.

\begin{corollary}[Geometric returns]
\label{cor:geometric}
Under the conditions of Theorem~\ref{thm:threshold} with $\psi(j)=\rho^{\,j-1}$, $0<\rho<1$,
\begin{equation}
\tau^\star=\min\left\{\Emax,\ \max\left\{1,\left\lceil\frac{\log\kappa}{\log\rho}\right\rceil\right\}\right\}.
\label{eq:kstar}
\end{equation}
For $\Emax\ge2$, the optimal rule stops after one epoch if and only if $\kappa\ge\rho$.
\end{corollary}

\begin{proof}
Here $m_{k+1}=\bar\ell\rho^{k}-\lambda\bar\delta$ is strictly decreasing, so Theorem~\ref{thm:threshold} applies, and $m_{k+1}\le0$ is equivalent to $\rho^k\le\kappa$, i.e.\ $k\ge\log\kappa/\log\rho$ since $\log\rho<0$. The compulsory first epoch and the cap give \eqref{eq:kstar}. Stopping at one is optimal iff $m_2=\bar\ell\rho-\lambda\bar\delta\le0$, i.e.\ $\kappa\ge\rho$.
\end{proof}

The epoch count \eqref{eq:kstar} decreases in the drift price $\lambda$ and the drift rate $\bar\delta$, and increases in the first-epoch progress $\bar\ell$ and the persistence $\rho$. The condition $\kappa\ge\rho$ states precisely when a client's second epoch is not worth taking: when the priced drift of an epoch exceeds the progress it is expected to deliver. If this holds for every client, \fedstop{} reduces to one-epoch training with step-normalised aggregation, which is optimal for the drift-priced utility rather than a failure of the rule. The same algebra calibrates the drift price: a client trains for exactly $k\ge2$ epochs when $\lambda\in[\bar\ell\rho^{k}/\bar\delta,\ \bar\ell\rho^{k-1}/\bar\delta)$, and $\lambda\ge\bar\ell\rho/\bar\delta$ forces one epoch. Measuring $\bar\ell$, $\bar\delta$ and $\rho$ in a pilot round therefore bounds the useful range of $\lambda$ before any tuning.

\begin{corollary}[Power-law returns]
\label{cor:power}
Under the conditions of Theorem~\ref{thm:threshold} with $\psi(j)=j^{-a}$, $a>0$,
\begin{equation}
\tau^\star=\min\bigl\{\Emax,\;\max\{1,\lceil\kappa^{-1/a}\rceil-1\}\bigr\}.
\label{eq:kstar_pow}
\end{equation}
For $\Emax\ge2$ it stops after one epoch if and only if $\kappa\ge2^{-a}$, and trains exactly $k\ge2$ epochs when $\lambda\in[\bar\ell(k+1)^{-a}/\bar\delta,\;\bar\ell k^{-a}/\bar\delta)$.
\end{corollary}

\begin{proof}
$\psi$ is decreasing, so Theorem~\ref{thm:threshold} applies, and $m_{k+1}=\bar\ell(k+1)^{-a}-\lambda\bar\delta\le0$ if and only if $k+1\ge\kappa^{-1/a}$.
\end{proof}

Under a power law, budgets grow faster as the drift price falls: polynomially in $1/\kappa$, against logarithmically for the geometric profile. A large exponent, however, makes the second epoch expensive. With the pilot value $a\approx2.4$ (Section~\ref{sec:experiments}), the rule stops after one epoch whenever $\kappa\ge2^{-2.4}\approx0.19$. This explains why the power-law rule chooses short budgets in Section~\ref{sec:model-A}.

Theorem~\ref{thm:threshold} asks for non-increasing net gains, but its proof uses only that the one-stage sets are nested. This matters when drift cost is not stationary, as Section~\ref{sec:model} finds.

\begin{corollary}[Declining drift cost, geometric returns]
\label{cor:declining}
Let $\E[\Delta\Lc(j)\mid\Fil_{j-1}]=\bar\ell\rho^{\,j-1}$ for $j\ge1$ and $\E[\delta(j)\mid\Fil_{j-1}]=\bar\delta\sigma^{\,j-2}$ for $j\ge2$, with $0<\rho<\sigma\le1$ and $\delta(1)$ arbitrary. Then the one-stage rule is optimal and
\begin{equation}
\tau^\star=\min\left\{\Emax,\ 1+\max\left\{0,\left\lceil\frac{\log(\kappa/\rho)}{\log(\rho/\sigma)}\right\rceil\right\}\right\},
\label{eq:kstar_decl}
\end{equation}
which reduces to \eqref{eq:kstar} when $\sigma=1$. The one-epoch condition remains $\kappa\ge\rho$, whatever $\sigma$ and the first-epoch drift. If instead $\sigma<\rho$, the one-stage sets need not be nested, and the one-stage rule can stop before a later epoch with positive net gain.
\end{corollary}

\begin{proof}
For $k\ge1$, $m_{k+1}=\bar\ell\rho^{k}-\lambda\bar\delta\sigma^{k-1}\le0$ if and only if $(\rho/\sigma)^{k-1}\le\kappa/\rho$. When $\rho<\sigma$ the left side decreases in $k$, so the sets $A_k=\{m_{k+1}\le0\}$ are deterministic and nested, and Lemma~\ref{lem:monotone} makes the one-stage rule optimal. Its first stopping epoch is $k=1$ if $\kappa\ge\rho$ and otherwise $1+\lceil\log(\kappa/\rho)/\log(\rho/\sigma)\rceil$. The cap gives \eqref{eq:kstar_decl}, and $1+\lceil\log(\kappa/\rho)/\log\rho\rceil=\lceil\log\kappa/\log\rho\rceil$ recovers \eqref{eq:kstar}. When $\sigma<\rho$ the left side increases, so $m_{k+1}\le0$ can hold for small $k$ and fail for larger $k$.
\end{proof}

Declining drift is therefore harmless as long as it declines more slowly than progress: the rule simply trains longer.

\begin{corollary}[Declining drift cost, power-law returns]
\label{cor:declining_pow}
Let $\E[\Delta\Lc(j)\mid\Fil_{j-1}]=\bar\ell j^{-a}$ for $j\ge1$, $a>0$, and $\E[\delta(j)\mid\Fil_{j-1}]=\bar\delta\sigma^{\,j-2}$ for $j\ge2$, with $0<\sigma\le1$ and $\delta(1)$ arbitrary. If $\sigma\ge(1-1/\Emax)^{a}$, the one-stage rule is optimal. It stops at the first $k$ with $(k+1)^{-a}\le\kappa\sigma^{k-1}$, capped at $\Emax$, and its one-epoch condition remains $\kappa\ge2^{-a}$.
\end{corollary}

\begin{proof}
For $k\ge1$, $m_{k+1}\le0$ if and only if $h(k)=(k+1)^{-a}\sigma^{1-k}\le\kappa$. The ratio $h(k+1)/h(k)=((k+1)/(k+2))^{a}/\sigma$ increases in $k$, so it is at most one for every $k\le\Emax-2$ exactly when $\sigma\ge(1-1/\Emax)^{a}$. The sets $\{m_{k+1}\le0\}$ are then nested, and Lemma~\ref{lem:monotone} applies. At $k=1$, $m_2\le0$ if and only if $\kappa\ge2^{-a}$.
\end{proof}

Unlike the geometric case, the condition tightens with the horizon, because polynomially decaying progress eventually declines more slowly than any geometric drift. For example, with $a=2.4$ and $\Emax=20$ it requires $\sigma\ge0.88$, and a drift ratio of $0.83$, as measured on FEMNIST, keeps the sets nested only up to twelve epochs.

\subsection{The cost of estimating the gains}\label{sec:rule-C}

A deployed client does not know $m_j$, so it estimates it. The following result gives the exact price of any deviation from $\tau^\star$ and bounds it for the plug-in rule used by \fedstop{}.

\begin{theorem}[Regret of estimated stopping rules]
\label{thm:regret}
Under the conditions of Theorem~\ref{thm:threshold}, every stopping time $\hat\tau$ satisfies
\begin{equation}
\begin{split}
\E[U(\tau^\star)]-\E[U(\hat\tau)]&=\sum_{j=2}^{\Emax}|m_j|\\
&\quad\times\Pr\big(\mathbf 1\{\hat\tau\ge j\}\ne\mathbf 1\{\tau^\star\ge j\}\big).
\end{split}
\label{eq:regret}
\end{equation}
In particular, let $\hat\tau=\min\{k:\hat m_{k+1}\le0\}\wedge\Emax$ for estimates $\hat m_{k+1}$ that are $\Fil_k$-measurable. Then
\begin{equation}
\begin{split}
\E[U(\tau^\star)]-\E[U(\hat\tau)]&\le\sum_{j=2}^{\Emax}|m_j|\\
&\quad\times\!\!\sum_{i=2}^{\min\{j,\tau^\star+1\}}\!\!\Pr\big(|\hat m_i-m_i|\ge|m_i|\big).
\end{split}
\label{eq:regretbound}
\end{equation}
\end{theorem}

\begin{proof}
Because $\{\tau\ge j\}\in\Fil_{j-1}$ and $m_j$ is deterministic, $\E[U(\tau)]=\sum_jm_j\Pr(\tau\ge j)$ for any stopping time. Since $\tau^\star$ is deterministic, the difference of the two expectations is $\sum_jm_j\big(\mathbf 1\{j\le\tau^\star\}-\Pr(\hat\tau\ge j)\big)$. For $2\le j\le\tau^\star$, $m_j>0$ and the term is $m_j\Pr(\hat\tau<j)$. For $j>\tau^\star$, $m_j\le0$ and the term is $|m_j|\Pr(\hat\tau\ge j)$. The $j=1$ term vanishes. This is \eqref{eq:regret}. For the plug-in rule, $\hat\tau\ge j$ if and only if $\hat m_i>0$ for all $2\le i\le j$. If $j\le\tau^\star$, a disagreement requires some $i\le j$ with $\hat m_i\le0<m_i$, hence $|\hat m_i-m_i|\ge|m_i|$. If $j>\tau^\star$, it requires $\hat m_{\tau^\star+1}>0\ge m_{\tau^\star+1}$, hence the same event at $i=\tau^\star+1$. A union bound gives \eqref{eq:regretbound}.
\end{proof}

Theorem~\ref{thm:regret} makes three properties of the plug-in rule precise. First, the regret is zero whenever every estimate has the correct sign, so the rule needs to estimate only whether an epoch is worth taking, not its value. Second, each wrong decision is charged the net gain $|m_j|$ of the epoch it adds or forgoes. These gains vanish near the threshold, where errors are most likely, so being off by an epoch near $\tau^\star$ costs little. Third, the event $|\hat m_i-m_i|\ge|m_i|$ compares estimation error with the margin $|m_i|$. Decisions far from the threshold, including the reduction to one epoch when $\kappa$ is well above $\rho$, are therefore robust to estimation noise.

\subsection{Correlated progress}\label{sec:rule-D}

Theorems~\ref{thm:threshold} and~\ref{thm:regret} take each $m_j$ to be deterministic, independent of how earlier epochs went. This assumption is not necessary: both results extend to trajectory-dependent gains, provided stopping decisions are nested.

\begin{proposition}[Trajectory-dependent gains]
\label{prop:trajectory}
If the sets $\{m_{k+1}\le0\}$, $k<\Emax$, are nested almost surely, then $\tau^\star$ in \eqref{eq:threshold} is an earliest optimal stopping time, and every stopping time $\hat\tau$ satisfies
\begin{equation}
\begin{split}
\E[U(\tau^\star)]-\E[U(\hat\tau)]&=\sum_{j=2}^{\Emax}\E\Big[|m_j|\\
&\quad\times\mathbf 1\big\{\mathbf 1\{\hat\tau\ge j\}\ne\mathbf 1\{\tau^\star\ge j\}\big\}\Big].
\end{split}
\label{eq:regret_traj}
\end{equation}
\end{proposition}

\begin{proof}
As noted before Corollary~\ref{cor:declining}, the proof of Theorem~\ref{thm:threshold} needs only nested sets $A_k=\{m_{k+1}\le0\}$. For \eqref{eq:regret_traj}, $\{\tau\ge j\}\in\Fil_{j-1}$ gives $\E[U(\tau)]=\E[g(1)]+\sum_{j\ge2}\E\big[m_j\mathbf 1\{\tau\ge j\}\big]$ for any stopping time. On every path, $m_j>0$ for $2\le j\le\tau^\star$ by the definition of $\tau^\star$, and $m_j\le0$ for $j>\tau^\star$ by nestedness. Each term of the difference therefore equals $|m_j|$ times the indicator of disagreement.
\end{proof}

Correlation therefore matters only if it can reverse a stopping decision. An autoregressive model of progress shows when it cannot.

\begin{corollary}[Correlated progress]
\label{cor:correlated}
Let $\Delta\Lc(j)=\bar\ell\psi(j)+\varepsilon_j$ for a decreasing profile $\psi$, with $\varepsilon_j=\phi\,\varepsilon_{j-1}+\eta_j$, $\varepsilon_0=0$, $\E[\eta_j\mid\Fil_{j-1}]=0$ and $|\varepsilon_j|\le B$, and let $\E[\delta(j)\mid\Fil_{j-1}]=\bar\delta$ for $j\ge2$. Let $\bar r=\max_{1\le k\le\Emax-2}\psi(k+2)/\psi(k+1)$, which is $\rho$ for the geometric and $(1-1/\Emax)^{a}$ for the power-law profile. Then $m_{k+1}=\bar\ell\psi(k+1)+\phi\varepsilon_k-\lambda\bar\delta$, and if
\begin{equation}
\lambda\bar\delta\,(1-\bar r)\ \ge\ |\phi|\,B\,(1+\bar r),
\label{eq:margin}
\end{equation}
the sets $\{m_{k+1}\le0\}$ are nested, so the rule \eqref{eq:threshold} with these $m_{k+1}$ is optimal.
\end{corollary}

\begin{proof}
Given $\Fil_k$, $\varepsilon_k$ is known and $\E[\varepsilon_{k+1}\mid\Fil_k]=\phi\varepsilon_k$, which gives $m_{k+1}$. If $m_{k+1}\le0$, then $\bar\ell\psi(k+1)\le\lambda\bar\delta+|\phi|B$, and since $\psi(k+2)\le\bar r\psi(k+1)$, $m_{k+2}\le\bar r(\lambda\bar\delta+|\phi|B)+|\phi|B-\lambda\bar\delta\le0$ by \eqref{eq:margin}. Proposition~\ref{prop:trajectory} then applies.
\end{proof}

Correlation is harmless when an epoch's drift cost dominates the correlated part of its progress. The optimal rule then corrects the expected gain by $\phi\varepsilon_k$, where $\varepsilon_k$ is the deviation of the last epoch's progress from its expected value, and it reduces to Corollaries~\ref{cor:geometric} and~\ref{cor:power} when $\phi=0$.

\subsection{From theory to a deployable rule}\label{sec:rule-E}

In deployment, the profile parameters are estimated online. Each participating client evaluates its full training loss at initialisation and after every executed epoch. These monitoring passes are counted as computation (Appendix). A single round rarely yields enough epochs to estimate a decay rate. Each client therefore estimates its profile shape ($\hat\rho_i$ or $\hat a_i$) from the per-epoch loss decreases of its current round and its last $w$ participations, the rounds in which it was sampled. The loss scale $\hat{\bar\ell}_i$ changes as the global model improves, so it is fitted within the round. The shape is assumed stable over nearby participations. The drift rate is a shrinkage mean, $\hat{\bar\delta}_i=(n_0\delta^{\mathrm{prior}}_i+\sum_{j\le k}\delta_i(j))/(n_0+k)$ with $n_0=2$. Its prior is the client's recent mean drift, or a population or pilot value before the client has history. Both terms include the first epoch. The shape is fitted by a penalised profile likelihood with a pilot prior, detailed in the appendix. The client continues while the plug-in look-ahead statistic
\begin{equation}
\hat m_i(k+1)=\hat{\bar\ell}_i\,\hat\psi_i(k+1)-\lambda\,\hat{\bar\delta}_i
\label{eq:plugin}
\end{equation}
is positive. Because clients return different numbers of steps $K_i=\tau_iS_i$, the server aggregates with step normalisation \eqref{eq:fednova}. Algorithm~\ref{alg:fedstop} summarises the method. Once these estimates are formed, each stopping decision takes $O(1)$ time. Refitting the shape scores $G$ candidate values over its search range ($G=200$ here) and refines the best by golden-section search, at a cost of $O(G(w+1)\Emax)$ scalar operations. Each client stores $O((w+1)\Emax)$ scalars. All of these costs are independent of model size. The dominant cost is evaluating the training loss, one forward pass over the local data, or a fixed subset of it, per epoch (Section~\ref{sec:consequences-C}).

\begin{figure}[t]
\centering
\includegraphics[width=\textwidth]{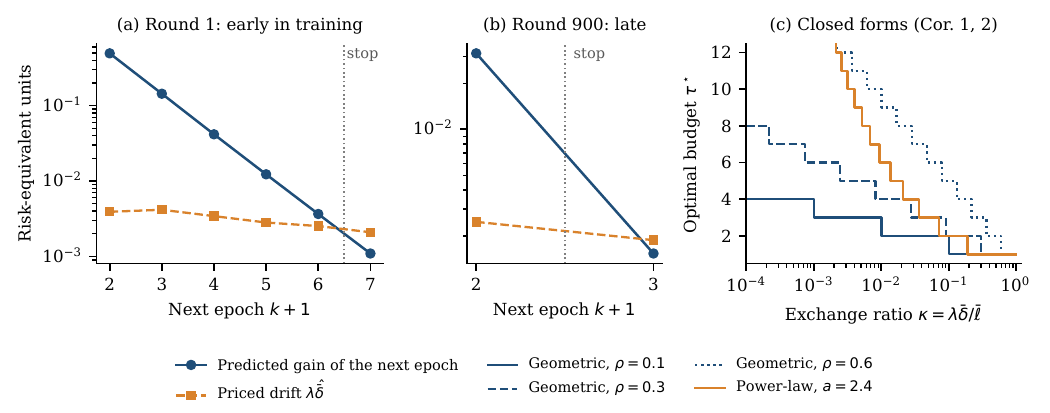}
\caption{How \fedstop{} decides. (a, b) Two client-rounds from the CIFAR-10 logs (geometric profile, seed 6): the predicted gain of the next epoch, $\hat{\bar\ell}_i\hat\psi_i(k+1)$, against its priced drift $\lambda\hat{\bar\delta}_i$. The client stops where the gain falls below the drift. (c) Optimal budgets from Corollaries~\ref{cor:geometric} and~\ref{cor:power} as functions of the exchange ratio $\kappa$.}
\label{fig:decision}
\end{figure}

Figure~\ref{fig:decision} shows the rule at work on two client-rounds from the CIFAR-10 logs. Early in training, the predicted gain of the next epoch starts near $0.5$, falls geometrically, and drops below the priced drift after six epochs. Late in training, it starts more than an order of magnitude lower and drops below the priced drift after two. In both cases the client stops at the crossing, as Theorem~\ref{thm:threshold} prescribes. The third panel gives the closed forms of Corollaries~\ref{cor:geometric} and~\ref{cor:power}. Budgets shrink as the exchange ratio $\kappa$ grows, logarithmically in $1/\kappa$ for the geometric profile and polynomially for the power law.

\begin{algorithm}[t]
\caption{\fedstop{}}
\label{alg:fedstop}
\begin{algorithmic}[1]
\Require rounds $T$, cap $\Emax$, drift price $\lambda$, profile $\psi$, window $w$
\For{$t=1,\dots,T$}
  \State broadcast $\bthe^{(t)}$ to sampled cohort $\mathcal S_t$
  \For{each client $i\in\mathcal S_t$ \textbf{in parallel}}
    \State $\bthe_i(0)\gets\bthe^{(t)}$, $k\gets0$
    \Repeat
      \State $k\gets k+1$ and run one local epoch
      \State observe $\Delta\Lc_i(k)$ and $\delta_i(k)$
      \State update $\hat\psi_i$ over the last $w$ participations, and $\hat{\bar\ell}_i,\hat{\bar\delta}_i$ in the round
      \State $\hat m\gets\hat{\bar\ell}_i\hat\psi_i(k+1)-\lambda\hat{\bar\delta}_i$
    \Until{$\hat m\le0$ \textbf{or} $k=\Emax$}
    \State $K_i\gets kS_i$ and return $\big(\bthe_i(k),K_i\big)$
  \EndFor
  \State $\bar K\gets\sum_{i\in\mathcal S_t}q_iK_i$
  \State $\bthe^{(t+1)}\gets\bthe^{(t)}+\bar K\sum_{i\in\mathcal S_t}q_i\big(\bthe_i(\tau_i)-\bthe^{(t)}\big)/K_i$
\EndFor
\end{algorithmic}
\end{algorithm}

\subsection{Predictions of the theory}\label{sec:rule-F}
\label{sec:predictions}

If the theory describes local training, five properties should be visible in training logs. We check each directly, independently of whether \fedstop{} improves accuracy. Together they form the evidence for the model.

\textbf{(P1) Geometric diminishing returns.} Within a round, the expected loss decrease per epoch decays geometrically, as expected under local linear convergence \citep{karimi2016pl,liu2022pl}.

\textbf{(P2) Additive, stationary drift cost.} Drift accrues epoch by epoch, $D_i(k)^2\approx\sum_{j\le k}\|v_j\|^2$, at a constant expected rate $\bar\delta$ over the decision-relevant epochs $j\ge2$. Corollaries~\ref{cor:declining} and~\ref{cor:declining_pow} relax this to a slowly declining rate.

\textbf{(P3) Training stage sets the budget.} By \eqref{eq:kstar}, a client's budget depends on the round only through $\kappa=\lambda\bar\delta/\bar\ell$ and $\rho$. As the global model improves, the progress $\bar\ell$ still available to a client shrinks, $\kappa$ grows, and budgets fall for all clients together.

\textbf{(P4) Clients with more data stop earlier.} An epoch of a client with more local examples takes more SGD steps. Under local linear convergence it therefore removes a larger share of its remaining loss. Its progress ratio $\rho$ is then smaller and, by \eqref{eq:kstar}, its budget shorter, other things equal.

\textbf{(P5) Nested stopping decisions.} Optimality needs only that a stopping decision, once taken, would not reverse (Proposition~\ref{prop:trajectory}). Correlated epoch gains preserve this under the margin \eqref{eq:margin}. The plug-in rule additionally treats gains as uncorrelated.

Whether budgets vary more across rounds (P3) or across clients (P4) is an empirical question that the model makes consequential. If round-level quantities dominate, permuting a round's budgets across its clients should leave training largely unchanged. We test this directly with a random-swap control.

\section{Experimental Setup}
\label{sec:experiments}

\textbf{Settings.} We use five main settings with 100 clients and 10\% uniform participation per round (Table~\ref{tab:settings}). They are CIFAR-10 with Dirichlet label partitions at $\alpha=0.1$ and $\alpha=0.5$ \citep{li2022niidbench}, CIFAR-100 at $\alpha=0.1$, and FEMNIST and Shakespeare with natural client identities \citep{caldas2018leaf}. Two further CIFAR-10 partitions, a near-IID Dirichlet split ($\alpha=1.0$) and a pathological two-class shard split \citep{mcmahan2017}, are used only for the progress-profile fits (marked $^\dagger$ in Table~\ref{tab:settings}). CIFAR uses a CNN with two convolutional and two dense layers, FEMNIST the LEAF CNN, and Shakespeare a two-layer LSTM. All methods use plain SGD without momentum or weight decay, batch size 50 (20 on FEMNIST), $\Emax=20$, and one client learning rate $\eta$ per setting. Shakespeare uses $\eta=1.47$, the value of the original FedAvg study \citep{mcmahan2017}, and clips gradients at norm 10. Every setting except Shakespeare holds out 10\% of each client's data for validation. Shakespeare splits each role's snippets in temporal order into 72\% training, 8\% validation and 20\% test data. On FEMNIST and Shakespeare, this hold-out also defines the cross-fitted tail described below.

\begin{table}[t]
\caption{Experimental settings ($\eta$: client learning rate, $E^\star$: tuned FedAvg epochs)}
\label{tab:settings}
\centering
\footnotesize
\setlength{\tabcolsep}{4pt}
\begin{tabular}{@{}llcccc@{}}
\toprule
Setting & Partition & Model & $\eta$ & $T$ & $E^\star$ \\
\midrule
CIFAR-10 & Dirichlet $\alpha=0.1$ & CNN & 0.03 & 1000 & 2 \\
CIFAR-10 & Dirichlet $\alpha=0.5$ & CNN & 0.03 & 1000 & 1 \\
CIFAR-10$^\dagger$ & Dirichlet $\alpha=1.0$ & CNN & 0.03 & 1000 & 1 \\
CIFAR-10$^\dagger$ & two-class shards & CNN & 0.03 & 1000 & 2 \\
CIFAR-100 & Dirichlet $\alpha=0.1$ & CNN & 0.03 & 500 & 2 \\
FEMNIST & natural writers & LEAF CNN & 0.03 & 1000 & 2 \\
Shakespeare & natural roles & LSTM & 1.47 & 500 & 1 \\
\bottomrule
\end{tabular}
\end{table}

\textbf{Baselines and controls.} All methods share each setting's learning rate. We compare with \textbf{FedAvg} \citep{mcmahan2017} at $E=1$, $E=10$ and the validation-tuned $E^\star\in\{1,2,5,10,20\}$. \textbf{FedNova} \citep{wang2020fednova} runs at $E^\star$ and rescales each client's update by its local step count. The adaptive aggregation interval of \textbf{Wang et al.}~(\citeyear{wang2019adaptive}) (their variant A) chooses one budget per round for all clients. On CIFAR-10 at $\alpha=0.1$, FEMNIST and Shakespeare we also run \textbf{per-client early stopping}, which stops a client once its local validation loss has not improved for a tuned number of epochs and returns its best iterate. We also run \textbf{FedProx} \citep{li2020fedprox}, our drift comparator. Its proximal term penalises the same squared distance from the global model that \fedstop{} prices, but during local training rather than at the decision to stop. Its proximal weight is selected from five values. On the same three settings we also run the client-side stopping rules \textbf{ALT} \citep{shenaj2025alt} and \textbf{FLASH} \citep{panchal2023flash}, described in Section~\ref{sec:consequences-D}.

\fedstop{}'s budgets vary both across rounds and across clients, which raises a question: which of the two drives its gains? A \emph{random-swap} control answers it. It reruns training with the same clients in each round, but hands that round's \fedstop{} budgets to its clients in random order. For example, if \fedstop{} gives clients A, B and C four, one and two epochs, the swap might give them one, two and four. Each round keeps the same budgets and the same total, but no client keeps its own. If matching budgets to clients matters, the swap should do worse than \fedstop{}. If only the amount of training per round matters, the two should perform alike. An \emph{equal-split} control goes further. It keeps each round's total but gives every client in the round the same share, which removes the budgets' heterogeneity altogether. Both controls replay the budgets of a \fedstop{} run, so they test its schedule retrospectively rather than act online. Clients train without loss monitoring, and the server aggregates with \eqref{eq:fednova}. Because clients differ in size, both controls execute slightly more SGD steps than \fedstop{}, about $5\%$ more on CIFAR-10 and under $2\%$ more on FEMNIST. We run the swap for the power-law profile in every setting and for the geometric profile wherever its budgets vary (CIFAR-10 at $\alpha=0.1$ and FEMNIST). We run the equal split for the geometric profile on the same two settings. Finally, a \emph{round-wise validation oracle} provides a reference point. It picks each round's budget using global validation loss, which no real federation can observe.

\textbf{Protocol.} Hyperparameters, including $E^\star$ and a separate drift price $\lambda$ for each profile, are selected on seeds 1--5 by mean validation accuracy over the last five evaluations. Test data are never used for selection. We report seeds 6--10, paired across methods by partition, initialisation and client sampling. Runs that diverge are retained and scored at chance accuracy for the remainder of the horizon. The window $w=4$ was fixed in advance to balance estimation noise against staleness, and spans about $40$ rounds at $10\%$ participation. Pooling over one or over all past participations does not do significantly better (Appendix). Each profile's drift price is the best of a five-value grid spanning the range that a pilot round calibrates (Section~\ref{sec:rule-B}). The selected prices range from $6.7\times10^{-4}$ to $1.27\times10^{4}$ (Appendix). The geometric shape is learned online. On CIFAR-10, however, it sits at its lower search bound in most decisions, so there the drift price effectively sets the budget. The power-law exponent mostly stays at its pilot value ($2.43$ on CIFAR-10, $2.48$ on Shakespeare), because a rule that seldom trains past one epoch seldom accumulates the rounds a refit needs. In seven of the ten setting--profile pairs the selected price is the grid's largest, which makes extra epochs expensive. In five of these pairs every budget is already one epoch, so a larger price could not change the rule. Only for the power-law rule on CIFAR-10 at $\alpha=0.1$ and on Shakespeare, where budgets still vary, could the optimum lie beyond the grid. Only geometric \fedstop{} on CIFAR-10 at $\alpha=0.1$ and both rules on FEMNIST select an interior price, and these are the settings where the geometric rule is active (Table~\ref{tab:spread}). Our analysis plan designated the power-law profile as primary and the geometric profile as secondary. We lead with the geometric profile because it fits local trajectories better (Section~\ref{sec:model}), and report both.

\textbf{Reproducibility.} Seeds 1--5 select every hyperparameter and seeds 6--10 report every result, so no reported number is used for selection. The appendix lists all grids, selected values, pilot priors and per-configuration results, and the code, configurations and training logs will be released with the paper.

\textbf{Epoch-level logs.} Runs log, for every client and epoch, the squared drift $D_i(k)^2$, its increment $\delta_i(k)$ and the squared per-epoch displacement $\|v_k\|^2$, together with each client's budget per round. \fedstop{} runs also log the local training loss after every epoch. For the drift checks we use the geometric random-swap runs: there, a client's budget comes from another client's decision, so which of its epochs are observed does not depend on its own trajectory. For the budget checks we use the budgets geometric \fedstop{} chose.

\textbf{Metrics.} Mean accuracy is top-1 test accuracy at round $T$. Worst-decile (tail) accuracy is the mean accuracy of the worst 10\% of the 100 clients. On CIFAR, where clients have no separate test sets, a client's accuracy is the per-class test accuracy reweighted by its training label histogram. On FEMNIST and Shakespeare we report two tails. In the \emph{cross-fitted} tail, each run's validation accuracies define its worst decile and test accuracy scores it. The \emph{shared-index} tail uses the ten client indices ranked lowest by tuned FedAvg on the selection seeds. On FEMNIST these indices identify different writers on different seeds. We give mean $\pm$ standard deviation over the five seeds and, for the principal contrasts, unadjusted paired 95\% $t$-intervals. For the headline tail gains over tuned FedAvg and Wang et al., and for the comparison with ALT and FLASH, we also report a Holm correction.

\section{Testing the Theory Against Local Training}
\label{sec:model}

Table~\ref{tab:scorecard} summarises the outcome: each element of the model, the result that relies on it, and the evidence for it. The subsections give the detail.

\subsection{Diminishing returns (P1)}\label{sec:model-A}

On held-out full-budget local trajectories, the geometric profile extrapolates better than the power-law profile for $75\%$, $73\%$, $79\%$ and $63\%$ of clients on the four CIFAR-10 partitions and $78\%$ on FEMNIST. This supports P1 and is why we lead with the geometric profile. Table~\ref{tab:spread} shows how each profile behaves as a stopping rule. Its interior fraction $P(1<\tau_i<\Emax)$ (Int.) is the share of client-rounds in which the budget lies strictly between the compulsory single epoch and the cap. It measures how often the rule is genuinely active, that is, how often the decision comes from the trade-off between progress and drift rather than from a boundary. The table also reports the mean budget $\bar\tau$ and its within-round variance (Var). The geometric rule is active in $98\%$ of decisions on CIFAR-10 at $\alpha=0.1$ and in $94\%$ on FEMNIST, where it adapts substantially. On CIFAR-10 at $\alpha=0.5$, CIFAR-100 and Shakespeare it is never active: the estimated $\kappa$ exceeds $\rho$ for every client, so the rule chooses one epoch, which Corollary~\ref{cor:geometric} identifies as optimal. The power-law rule mostly stops after one epoch, with mean budgets of $1.0$--$1.4$ outside FEMNIST, and accordingly gives a smaller tail gain than the geometric rule (Table~\ref{tab:leaf}).

\begin{table}[t]
\caption{Stopping behaviour of the two profiles}
\label{tab:spread}
\centering
\footnotesize
\setlength{\tabcolsep}{3pt}
\begin{tabular}{@{}lcccccc@{}}
\toprule
& \multicolumn{3}{c}{Geometric} & \multicolumn{3}{c}{Power-law} \\
\cmidrule(lr){2-4}\cmidrule(lr){5-7}
Setting & Int. & $\bar\tau$ & Var & Int. & $\bar\tau$ & Var \\
\midrule
CIFAR-10, $\alpha=0.1$ & 0.98 & 2.5 & 0.57 & 0.31 & 1.3 & 0.20 \\
CIFAR-10, $\alpha=0.5$ & 0.00 & 1.0 & 0.00 & 0.04 & 1.0 & 0.03 \\
CIFAR-100 & 0.00 & 1.0 & 0.00 & 0.00 & 1.0 & 0.00 \\
FEMNIST & 0.94 & 5.4 & 8.17 & 0.85 & 2.9 & 3.05 \\
Shakespeare & 0.00 & 1.0 & 0.00 & 0.25 & 1.4 & 1.28 \\
\bottomrule
\end{tabular}
\end{table}

\subsection{Drift cost (P2)}\label{sec:model-B}

\begin{figure}[t]
\centering
\includegraphics[width=0.65\textwidth]{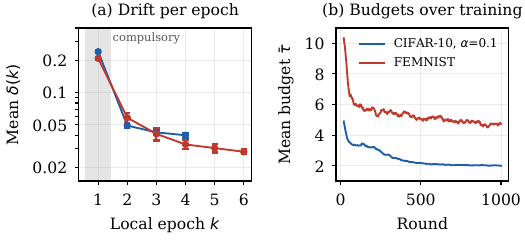}
\caption{(a) Mean drift increment $\delta(k)$ by local epoch within the same client-rounds of the geometric random-swap runs (budgets $\ge4$ on CIFAR-10 and $\ge6$ on FEMNIST, with 95\% intervals). (b) Mean geometric \fedstop{} budget per round (five seeds, 20-round moving average).}
\label{fig:model}
\end{figure}

Fig.~\ref{fig:model}(a) follows drift epoch by epoch within the same client-rounds. The first epoch moves the model $3.6$ (FEMNIST) to $4.9$ (CIFAR-10) times further than the second. On CIFAR-10, $\delta=0.240$ against $0.049$, $0.043$ and $0.040$ for epochs 2--4. On FEMNIST, $\delta=0.208$ against $0.058$, falling to $0.028$ by epoch 6. A stationary cost over all epochs would therefore be misspecified, but the anomaly sits entirely in the compulsory epoch, which never enters the optimal stopping decision (Corollary~\ref{cor:declining}). The deployed plug-in does include it (Section~\ref{sec:rule-E}), pricing continuation at a median 16 (CIFAR-10) and 7 (FEMNIST) times the drift later epochs realise. This shortens budgets, the safe direction, though only $2.1\%$ and $5.1\%$ of client-rounds stop after one epoch. After the first epoch, drift is close to stationary on CIFAR-10, with epochs 2--4 within $20\%$ of each other, whereas on FEMNIST it halves between epochs 2 and 6. Corollary~\ref{cor:declining} states when this matters. Fitting a geometric decay to epochs 2--6 gives $\sigma\approx0.83$ per epoch on FEMNIST ($0.90$ on CIFAR-10, epochs 2--4). The threshold rule therefore remains optimal provided per-epoch progress decays at least as fast, $\rho\le0.83$. The epoch logs confirm it: successive loss decreases over epochs 2--4 have median ratio $0.34$, and $86\%$ of them lie at or below $0.83$. Drift is additive: per-epoch squared displacements sum to a median $0.96$ (CIFAR-10) and $0.94$ (FEMNIST) of the final squared drift, and $\delta$ is negative in only $2.5\%$ and $5.1\%$ of epochs.

\subsection{Determinants of client budgets (P3, P4)}\label{sec:model-C}

Budgets fall as training proceeds, as P3 predicts (Fig.~\ref{fig:model}(b)). The mean geometric budget drops from $3.8$ epochs in the first hundred rounds to $2.0$ in the last hundred on CIFAR-10, and from $7.4$ to $4.7$ on FEMNIST. They fall faster than predicted, however. The fitted slope is $1.6$ (FEMNIST) to $4$ (CIFAR-10) times steeper than the slope implied by separately estimated within-round decay and across-round decay of first-epoch progress. The model therefore captures the direction of annealing but not its rate. To separate training stage from client identity, we fit each seed's budgets by least squares with additive round and client effects and average the explained variance over seeds. The round alone explains $44\%$ of budget variance on CIFAR-10 and $18\%$ on FEMNIST. Adding client identity raises this by only about ten points on both. The remaining within-round variation is therefore largely not a stable client property, but the round-to-round evolution of each client's estimates. The stable client component that does exist has the sign P4 predicts: budgets correlate negatively with the number of local examples (Spearman $-0.17$ on CIFAR-10 and $-0.09$ on FEMNIST, averaged over seeds).

\textbf{Swapped budgets.} Table~\ref{tab:swap} asks where \fedstop{}'s gains come from: from giving each client the right budget, or from giving each round the right amount of local work. A random swap keeps every round's budgets but breaks their link to clients, so it removes only the first. For both profiles and in every setting, no difference excludes zero. The point estimates even favour the swap slightly, which also runs $5.0\%$ (CIFAR-10) and $1.6\%$ (FEMNIST) more SGD steps because larger clients sometimes receive larger budgets. With five seeds the tail intervals are wide, so small effects cannot be excluded, and the largest lean, on the power-law rule's CIFAR-10 tail, stops just short of significance. The equal split, which removes heterogeneity altogether, also matches geometric \fedstop{} on both settings, and \fedstop{} has the higher tail on only one of five seeds in each. Neither control establishes equivalence, but both point the same way. Which client receives which budget therefore matters little, and on CIFAR-10 the two runs track each other throughout training (Fig.~\ref{fig:dynamics}). How much each round trains matters a great deal: holding each client's power-law budget at its average over the run, with the same epochs, lowers the CIFAR-10 tail by $3.9$ points ($[2.6,5.3]$). \fedstop{}'s value thus lies in the per-round schedule it derives, which both controls can only copy from it.

\subsection{Nested decisions (P5)}\label{sec:model-D}

Successive epoch gains are negatively autocorrelated. For the power-law runs, lag-one correlations of the residual per-epoch loss decreases average $-0.36$ on CIFAR-10 and $-0.50$ on Shakespeare. For the geometric runs, they average $-0.40$ on CIFAR-10 and $-0.32$ on FEMNIST. On the geometric runs the margin of Corollary~\ref{cor:correlated} fails in every client-round with at least four epochs. Under the fitted model, a stopping decision would reverse in $19\%$ (CIFAR-10) and $27\%$ (FEMNIST) of them. These are in-sample estimates on short, self-stopped trajectories, which fit each path's own scale and omit unobserved continuations, so they are indicative only. Little is lost in practice: a rule that solves the autoregressive stopping problem exactly by dynamic programming matches the one-stage rule that uses the same loss-scale estimator (Table~\ref{tab:scorecard}).

\begin{table}[t]
\caption{Model assumptions and their empirical support}
\label{tab:scorecard}
\centering
\footnotesize
\setlength{\tabcolsep}{2.5pt}
\renewcommand{\arraystretch}{1.1}
\begin{tabular}{@{}>{\raggedright\arraybackslash}p{0.33\linewidth}>{\raggedright\arraybackslash}p{0.17\linewidth}>{\raggedright\arraybackslash}p{0.46\linewidth}@{}}
\toprule
Model element & Used in & Evidence (CIFAR-10, FEMNIST) \\
\midrule
Geometric returns (P1) & Cor.~\ref{cor:geometric} & preferred for 63--79\% of clients \\
Additive drift, stationary or slowly declining (P2) & \eqref{eq:utility}, Cor.~\ref{cor:geometric},~\ref{cor:declining} & ratio $0.94$--$0.96$, stationary on CIFAR-10, $\rho\le\sigma$ in $86\%$ of FEMNIST epochs \\
Budgets fall in training, and annealing matters (P3) & \eqref{eq:kstar} & $3.8\!\to\!2.0$ and $7.4\!\to\!4.7$ epochs, and fixing budgets per client costs $3.9$ tail points \\
Round, not client, sets the budget (P3) & \eqref{eq:kstar} & random swaps match \fedstop{} (Table~\ref{tab:swap}) \\
Over-pricing is safe & Cor.~\ref{cor:geometric} & no failure at the two largest $\lambda$ \\
One-stage rule loses little to correlation & Prop.~\ref{prop:trajectory} & exact AR(1) rule, same scale: $+0.07$ points, $[-0.57,0.72]$ \\
\bottomrule
\end{tabular}
\end{table}

\begin{table}[t]
\caption{Top-1 accuracy (\%, five seeds)}
\label{tab:main}
\centering
\footnotesize
\setlength{\tabcolsep}{5pt}
\begin{tabular}{lccccc}
\toprule
& CIFAR-10 & CIFAR-10 & CIFAR-100 & & \\
Method & $\alpha=0.1$ & $\alpha=0.5$ & $\alpha=0.1$ & FEMNIST & Shakespeare \\
\midrule
FedAvg, $E=1$ & 60.7\,$\pm$\,1.3 & 67.8\,$\pm$\,1.0 & 29.0\,$\pm$\,1.1 & 81.7\,$\pm$\,1.9 & 50.0\,$\pm$\,0.4 \\
FedAvg, $E=10$ & 60.5\,$\pm$\,0.7 & 51.2\,$\pm$\,23.1 & 14.2\,$\pm$\,12.0 & 81.3\,$\pm$\,1.3 & 47.5\,$\pm$\,0.4 \\
FedAvg, tuned $E^\star$ & 60.2\,$\pm$\,4.0 & 67.8\,$\pm$\,1.0 & 30.4\,$\pm$\,1.8 & 81.8\,$\pm$\,1.4 & 50.0\,$\pm$\,0.4 \\
\citet{wang2019adaptive} & 62.0\,$\pm$\,1.1 & 67.8\,$\pm$\,1.2 & 30.1\,$\pm$\,1.8 & 82.0\,$\pm$\,1.5 & 50.0\,$\pm$\,0.4 \\
FedNova \citep{wang2020fednova} & 63.1\,$\pm$\,2.2 & 68.3\,$\pm$\,0.6 & \textbf{30.7\,$\pm$\,0.9} & \textbf{82.1\,$\pm$\,1.3} & 50.1\,$\pm$\,0.5 \\
\rowcolor{fsshade}\fedstop{} (power-law) & 62.8\,$\pm$\,1.6 & \textbf{68.5\,$\pm$\,0.6} & 29.6\,$\pm$\,0.6 & 81.8\,$\pm$\,1.6 & 48.2\,$\pm$\,0.5 \\
\rowcolor{fsshade}\fedstop{} (geometric) & \textbf{63.5\,$\pm$\,0.7} & 68.0\,$\pm$\,0.7 & 29.5\,$\pm$\,0.9 & 81.1\,$\pm$\,1.9 & \textbf{50.1\,$\pm$\,0.4} \\
\midrule
\multicolumn{6}{@{}l}{\textit{Reference, not deployable}} \\
Round-wise validation oracle & 64.0\,$\pm$\,0.5 & 66.7\,$\pm$\,0.8 & 29.3\,$\pm$\,0.4 & 82.6\,$\pm$\,1.0 & 49.4\,$\pm$\,0.5 \\
\bottomrule
\end{tabular}
\end{table}

\section{Performance in Federated Training}
\label{sec:consequences}

\subsection{A fixed budget is fragile}\label{sec:consequences-A}

\begin{figure}[t]
\centering
\includegraphics[width=0.65\textwidth]{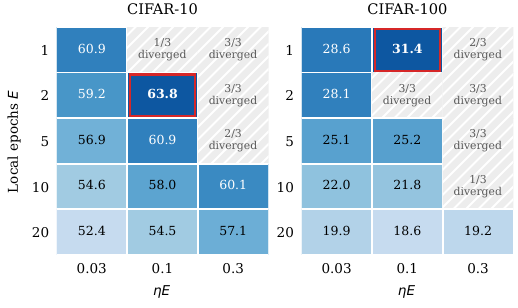}
\caption{Validation accuracy (\%) of fixed-budget FedAvg over epochs $E$ and exposure $\eta E$, three seeds per cell. Hatched cells diverged on the stated number of seeds, and the red box marks the best configuration.}
\label{fig:grid}
\end{figure}

Fig.~\ref{fig:grid} reports a separate grid of 90 FedAvg runs on CIFAR-10 and CIFAR-100, crossing $E\in\{1,2,5,10,20\}$ with three values of the per-round exposure $\eta E$, on three seeds per cell. The best fixed configuration is two epochs at $\eta E=0.1$ on CIFAR-10 ($63.8\%$) but one epoch at $\eta E=0.1$ on CIFAR-100 ($31.4\%$), where the same two-epoch configuration diverges on every seed. Across both datasets, nine of the thirty cells diverge on at least one seed. At the main learning rate (Table~\ref{tab:settings}), the tuned CIFAR-100 budget is instead two epochs, so a global budget must be retuned for each dataset and learning rate. \fedstop{} still has one tuned scalar, the drift price, but within that price the budget responds to observed progress.

\subsection{Accuracy and worst-client performance}\label{sec:consequences-B}

Table~\ref{tab:main} reports top-1 accuracy in all five settings. Table~\ref{tab:leaf} reports worst-decile accuracy where it is informative: on CIFAR-10 at $\alpha=0.1$, the CIFAR partition where geometric budgets vary, and on the naturally partitioned FEMNIST and Shakespeare. Both list the methods run in every setting, and bold marks the best by unrounded mean, excluding the oracle. Across the five settings, geometric \fedstop{} differs from validation-tuned FedAvg by $+3.3$, $+0.2$, $-0.9$, $-0.7$ and $+0.1$ points in mean accuracy, never trailing by more than one. On CIFAR-10 at $\alpha=0.1$, the budget selected on seeds 1--5 ($E^\star=2$) transfers poorly to seeds 6--10. It scores below both $E=1$ and $E=10$, with the widest spread in its column. Where the rule is active, it improves the worst decile. On CIFAR-10 its tail of $39.0\%$ exceeds tuned FedAvg by $8.25$ points (paired 95\% CI $[5.77,10.74]$), FedNova by $4.22$ ($[0.15,8.29]$) and Wang et al. by $6.29$ ($[4.39,8.18]$). The FedNova interval excludes zero only narrowly. On FEMNIST its cross-fitted tail exceeds tuned FedAvg by $4.84$ points ($[2.06,7.62]$) and Wang et al. by $4.37$ ($[2.97,5.77]$), but not FedNova ($+1.53$, $[-0.68,3.75]$). The four tail gains over tuned FedAvg and Wang et al. on CIFAR-10 and FEMNIST all remain significant after a Holm correction (largest adjusted $p=0.008$). The tables report the final round, the prespecified endpoint. Averaged over the last 50 rounds instead, the tail gains over tuned FedAvg are smaller, $5.0$ points on CIFAR-10 ($[3.6,6.3]$) and $3.0$ on FEMNIST ($[2.2,3.9]$), but they remain significant and positive on every seed. The shared-index tail is a weaker check, because its clients are not the worst decile of the runs. On it, the gain over tuned FedAvg is $1.55$ points ($[0.12,2.98]$) and the difference from FedNova is negligible. In the other three settings the geometric rule returns one epoch for every client and performs like one-epoch FedAvg. Forcing a second epoch there costs $1.8$ points on Shakespeare ($[1.6,1.9]$, every seed).

\begin{table}[t]
\caption{Worst-decile accuracy (\%, five seeds)}
\label{tab:leaf}
\centering
\footnotesize
\setlength{\tabcolsep}{2pt}
\begin{tabular}{lccccc}
\toprule
& CIFAR-10 & \multicolumn{2}{c}{FEMNIST} & \multicolumn{2}{c}{Shakespeare} \\
\cmidrule(lr){3-4}\cmidrule(lr){5-6}
Method & $\alpha=0.1$ & Cross-fit. & Shared & Cross-fit. & Shared \\
\midrule
FedAvg, $E=1$ & 28.7\,$\pm$\,3.7 & 67.6\,$\pm$\,6.3 & 75.1\,$\pm$\,3.1 & 49.4\,$\pm$\,1.7 & 50.7\,$\pm$\,0.5 \\
FedAvg, $E=10$ & 35.6\,$\pm$\,4.1 & 69.6\,$\pm$\,5.8 & 76.3\,$\pm$\,2.6 & 47.2\,$\pm$\,1.6 & 47.9\,$\pm$\,0.8 \\
FedAvg, tuned $E^\star$ & 30.7\,$\pm$\,2.0 & 68.3\,$\pm$\,5.3 & 75.5\,$\pm$\,3.9 & 49.4\,$\pm$\,1.7 & 50.7\,$\pm$\,0.5 \\
\citet{wang2019adaptive} & 32.7\,$\pm$\,3.0 & 68.8\,$\pm$\,5.8 & 75.2\,$\pm$\,3.9 & 50.0\,$\pm$\,1.4 & 50.5\,$\pm$\,0.6 \\
FedNova & 34.8\,$\pm$\,3.4 & 71.6\,$\pm$\,6.3 & 77.0\,$\pm$\,3.7 & \textbf{50.4\,$\pm$\,1.9} & 50.7\,$\pm$\,0.5 \\
\rowcolor{fsshade}\fedstop{} (power-law) & 33.8\,$\pm$\,4.6 & 71.5\,$\pm$\,4.7 & 76.5\,$\pm$\,4.9 & 46.8\,$\pm$\,1.0 & 48.6\,$\pm$\,1.0 \\
\rowcolor{fsshade}\fedstop{} (geometric) & \textbf{39.0\,$\pm$\,2.3} & \textbf{73.1\,$\pm$\,5.2} & \textbf{77.0\,$\pm$\,3.6} & 49.5\,$\pm$\,1.4 & \textbf{50.8\,$\pm$\,0.7} \\
\midrule
\multicolumn{6}{@{}l}{\textit{Reference, not deployable}} \\
Round-wise validation oracle & 37.8\,$\pm$\,6.5 & 73.9\,$\pm$\,5.7 & 77.9\,$\pm$\,1.5 & 48.7\,$\pm$\,1.8 & 50.0\,$\pm$\,0.3 \\
\bottomrule
\end{tabular}
\end{table}

\begin{table}[t]
\caption{\fedstop{} against controls built from its own budgets: every difference is within noise (\%, five seeds, $\Delta$: \fedstop{} minus control with 95\% CI)}
\label{tab:swap}
\centering
\scriptsize
\setlength{\tabcolsep}{3pt}
\begin{tabular}{lllcccccc}
\toprule
& & & \multicolumn{3}{c}{Top-1 accuracy} & \multicolumn{3}{c}{Worst-decile accuracy} \\
\cmidrule(lr){4-6}\cmidrule(lr){7-9}
Profile & Setting & Control & \fedstop{} & Control & $\Delta$ & \fedstop{} & Control & $\Delta$ \\
\midrule
Geometric & CIFAR-10, $\alpha=0.1$ & Random swap & 63.5 & 63.6 & $-0.06$ [$-1.33$, $1.22$] & 39.0 & 39.6 & $-0.61$ [$-2.46$, $1.24$] \\
Geometric & CIFAR-10, $\alpha=0.1$ & Equal split & 63.5 & 63.5 & $+0.01$ [$-0.26$, $0.28$] & 39.0 & 39.3 & $-0.35$ [$-2.09$, $1.39$] \\
Geometric & FEMNIST & Random swap & 81.1 & 81.5 & $-0.46$ [$-1.50$, $0.58$] & 73.1 & 76.5 & $-3.33$ [$-7.87$, $1.22$] \\
Geometric & FEMNIST & Equal split & 81.1 & 81.7 & $-0.64$ [$-2.49$, $1.21$] & 73.1 & 74.4 & $-1.21$ [$-3.76$, $1.34$] \\
Power-law & CIFAR-10, $\alpha=0.1$ & Random swap & 62.8 & 63.9 & $-1.11$ [$-2.24$, $0.03$] & 33.8 & 36.3 & $-2.54$ [$-5.19$, $0.11$] \\
Power-law & FEMNIST & Random swap & 81.8 & 82.2 & $-0.45$ [$-1.04$, $0.15$] & 71.5 & 71.3 & $+0.20$ [$-8.63$, $9.03$] \\
Power-law & Shakespeare & Random swap & 48.2 & 48.1 & $+0.09$ [$-0.26$, $0.44$] & 46.8 & 47.0 & $-0.19$ [$-1.86$, $1.47$] \\
\bottomrule
\end{tabular}
\end{table}

The gain is also stable across seeds (standard deviations in Table~\ref{tab:main}). On CIFAR-10, \fedstop{}'s worst seed has a tail of $36.6\%$, above the average of FedNova ($34.8\%$) and Wang et al.\ ($32.7\%$). Its mean accuracy varies a third as much as FedNova's and beats Wang et al. on every seed ($+1.58$ points, $[0.57,2.58]$). The tail gain is not explained by training volume. FedAvg with $E=10$ runs four times as many epochs as geometric \fedstop{} on CIFAR-10 but has a lower tail ($35.6\%$). On FEMNIST it runs nearly twice as many epochs for a cross-fitted tail of $69.6\%$ against $73.1\%$. Nor does it depend on which client receives which budget, since a random swap or an equal split of its budgets matches it (Section~\ref{sec:model-C}). What remains is the per-round schedule: step-normalised budgets that shrink as training proceeds, whose level matters more than how it is divided among clients.

\begin{figure}[t]
\centering
\includegraphics[width=0.8\textwidth]{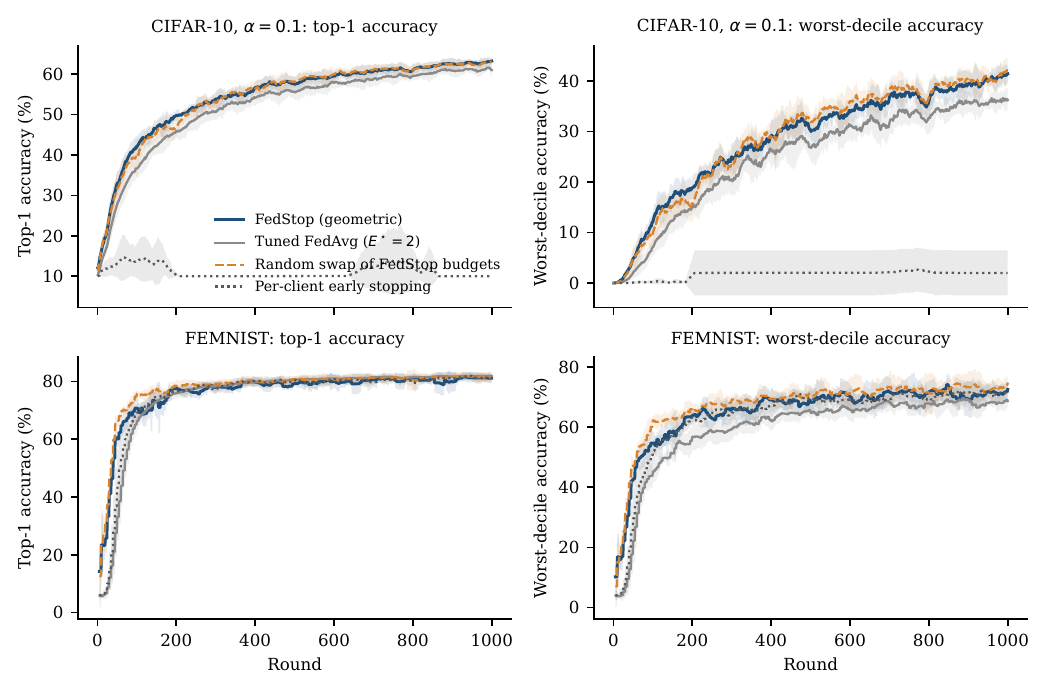}
\caption{Accuracy over training on seeds 6--10 (mean $\pm$ one standard deviation, 25-round moving average). The tables report the final round. On CIFAR-10, early stopping finishes at chance accuracy.}
\label{fig:dynamics}
\end{figure}

\textbf{Over training.} Fig.~\ref{fig:dynamics} shows that the tail advantage over tuned FedAvg opens early and persists. By round 250 the smoothed worst decile is $21.7\%$ against $18.5\%$ on CIFAR-10 and $62.8\%$ against $58.3\%$ on FEMNIST, and the gap holds to the end of training.

\begin{figure}[t]
\centering
\includegraphics[width=0.8\textwidth]{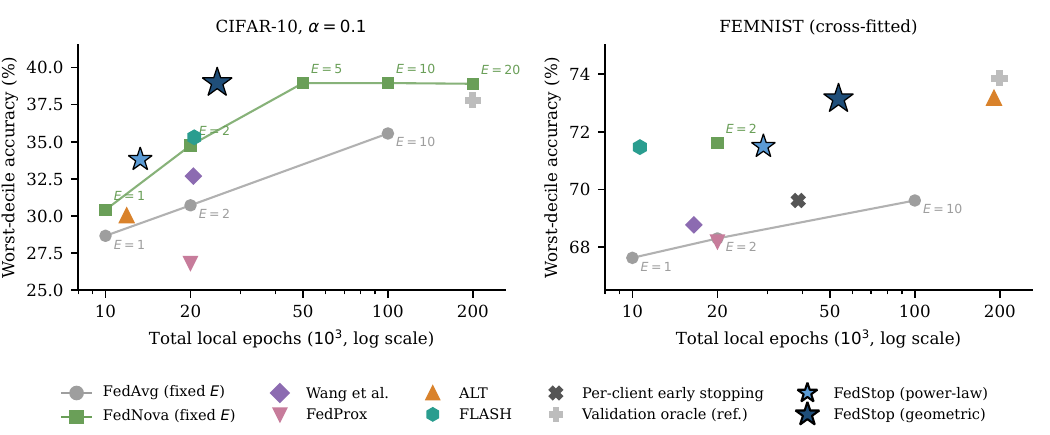}
\caption{Worst-decile accuracy against total local epochs (means over the five seeds). Lines trace fixed-budget sweeps of FedAvg and FedNova, and \fedstop{} reaches a higher tail with fewer epochs. Per-client early stopping on CIFAR-10 ends at chance and lies below the axis.}
\label{fig:efficiency}
\end{figure}

\textbf{Epoch efficiency.} Fig.~\ref{fig:efficiency} places every method by its total local epochs and its worst-decile accuracy. On CIFAR-10, geometric \fedstop{} reaches a $39.0\%$ tail with $24.9$k local epochs. No fixed budget matches it as cheaply. FedNova needs two to eight times as many epochs to reach the same tail, at $E=5$, $10$ or $20$, and then loses $0.9$--$4.3$ points of mean accuracy, every interval excluding zero (Appendix). Every FedAvg budget stays below it, and even the oracle reaches only $37.8\%$ with $200$k epochs. The client-side rules ALT and FLASH use fewer epochs on CIFAR-10 but reach clearly lower tails. On FEMNIST, ALT is the only deployable method whose tail is as high as \fedstop{}'s, and it needs $3.6$ times as many epochs. FLASH and FedNova come within two points with fewer epochs, a difference within noise. \fedstop{} comes close to the oracle's tail there at about a quarter of its epochs. \fedstop{} thus spends local epochs where they raise the accuracy of the worst-off clients.

\textbf{Robustness.} \fedstop{} is robust where a heuristic is not. Per-client early stopping ended at chance accuracy on all five CIFAR-10 seeds. It never learned on two of them and fell back after peaks of $27$--$43\%$ on the other three (Fig.~\ref{fig:dynamics}), and it collapsed at every patience value on some selection seed. On Shakespeare it trained $2.2$ times as many epochs for $0.75$ points less accuracy on every seed ($49.4\%$, $[0.53,0.96]$). On FEMNIST the two are level ($80.7\%$ and a cross-fitted tail of $69.6\%$, against \fedstop{} in Tables~\ref{tab:main} and~\ref{tab:leaf}). \fedstop{} itself failed only at drift prices low enough to train two to five times longer than the selected one, and never at the two largest. An upward mis-set is therefore safe in practice as well as in principle. A drift price transferred from another setting also stays level with the locally tuned one on FEMNIST and Shakespeare. The one-stage rule is also more robust than the sequential alternatives we prespecified. A sequential probability ratio test and an upper-confidence-bound rule trained far longer and collapsed on five and four of the five CIFAR-10 seeds, while the exact autoregressive rule matched the one-stage rule (Appendix).

\textbf{Other mechanisms.} FedProx, which penalises the drift that \fedstop{} prices, collapsed on one CIFAR-10 seed ($51.6\%$ over the five, tail $26.8\%$). It reaches $82.0\%$ on FEMNIST (cross-fitted tail $68.2\%$) and is the most accurate method on Shakespeare ($50.7\%$, tail $49.2\%$, against \fedstop{} in Tables~\ref{tab:main} and~\ref{tab:leaf}). The two are complementary rather than competing: FedProx changes the local objective, \fedstop{} its length.  Option C of Wang et al., which trades local computation against aggregation cost, and SCAFFOLD, which corrects the direction of local updates, are also reported in the appendix. Option C with its original $\phi$ is level with \fedstop{} on CIFAR-10.

\subsection{Controlling the cost of loss monitoring}\label{sec:consequences-C}

Beyond training, \fedstop{} adds two computations: a constant-time decision per epoch (Section~\ref{sec:rule-E}) and loss monitoring, a forward pass that measures each epoch's loss decrease. Monitoring cannot simply reuse the losses recorded during training, which are measured on different minibatches while the model moves, because the decision depends on small differences between consecutive losses. Its cost, however, is controllable by design. Evaluating the loss on a fixed subset of $\min(128,n_i)$ local examples, the same examples at every evaluation so that differences stay precise, bounds monitoring by a constant per epoch, independent of the local dataset size. We measured this variant on CIFAR-10 with both profiles, on CIFAR-100 and on FEMNIST, keeping every selected drift price. It lowers the modeled work in every setting. With it, the power-law rule on CIFAR-10 and the geometric rule on CIFAR-100 reach the accuracy target with less work than two-epoch FedAvg. Where the geometric rule is active, it still uses more work than tuned FedAvg. That extra work is local training, which buys the tail gain more cheaply than any fixed budget (Fig.~\ref{fig:efficiency}). On FEMNIST, however, FLASH reaches a statistically indistinguishable tail with about a fifth of the epochs. No mean-accuracy difference from full monitoring excludes zero. On CIFAR-10, the geometric rule's tail fell by $1.3$ points ($[0.1,2.6]$) in the primary run. A prespecified re-selection then chose the same drift price, and its repeat on the same seeds shows no tail difference from full monitoring ($-0.2$, $[-2.5,2.1]$), although it is not an independent replication. The appendix reports the measured work.

\subsection{Client-side stopping rules}\label{sec:consequences-D}

ALT and FLASH also set a budget for each client, so they test whether \fedstop{}'s model-based rule improves on simpler signals. We port each rule's local stopping criterion into our training protocol, with the same learning rates, seeds and step-normalised aggregation. These are ports of the stopping rules, not reproductions of the authors' complete systems. ALT stops a client once the cosine similarity between its local features and the global model's features on the same batch falls below a threshold $\tau$. The features are the first dense pre-activation for the CNNs and the final LSTM hidden state for Shakespeare. FLASH stops once an epoch reduces the mean validation cross-entropy by less than $\gamma$. A five-epoch FedAvg pilot sets each grid: the 10th to 90th percentiles of the pilot's feature similarity for ALT, and 0.1 to 3 times its median first-epoch validation-loss drop for FLASH. Each threshold is then selected on seeds 1--5 like every other hyperparameter, with up to two outward extensions of the grid. Three selections, ALT on CIFAR-10 and Shakespeare and FLASH on Shakespeare, remain at the edge of the grid after both extensions. They are therefore bounded-search winners rather than interior optima.

Table~\ref{tab:clientrules} reports the results, with bold marking the best method in each column. On CIFAR-10, geometric \fedstop{} has the highest worst-decile accuracy, $8.95$ points above ALT ($[1.40,16.50]$) and $3.68$ above FLASH ($[1.52,5.83]$), and it is ahead on every seed. These unadjusted intervals exclude zero, but neither difference survives a Holm correction over the six comparator tails (adjusted $p=0.15$ and $0.055$). Mean accuracy is level with both rules. On FEMNIST and Shakespeare every tail interval includes zero, so the three rules are level there. They reach these levels with very different budgets. On FEMNIST, ALT's similarity signal rarely fires, so it trains $19.1$ epochs per participation, close to the cap, whereas FLASH stops after about one epoch and \fedstop{} after $5.4$. On CIFAR-10 and Shakespeare, \fedstop{} uses more modeled work than both rules (Appendix).

\begin{table}[t]
\caption{Client-side stopping rules (\%, five seeds)}
\label{tab:clientrules}
\centering
\scriptsize
\setlength{\tabcolsep}{2.5pt}
\begin{tabular}{lcccccccc}
\toprule
& \multicolumn{2}{c}{CIFAR-10, $\alpha=0.1$} & \multicolumn{3}{c}{FEMNIST} & \multicolumn{3}{c}{Shakespeare} \\
\cmidrule(lr){2-3}\cmidrule(lr){4-6}\cmidrule(lr){7-9}
Method & Top-1 & Worst decile & Top-1 & Cross-fit. & Shared & Top-1 & Cross-fit. & Shared \\
\midrule
ALT \citep{shenaj2025alt} & 61.4\,$\pm$\,2.1 & 30.0\,$\pm$\,7.2 & 81.6\,$\pm$\,1.2 & \textbf{73.2\,$\pm$\,3.2} & \textbf{78.8\,$\pm$\,3.0} & 50.0\,$\pm$\,0.5 & 49.7\,$\pm$\,1.0 & 50.4\,$\pm$\,0.5 \\
FLASH \citep{panchal2023flash} & 63.2\,$\pm$\,1.4 & 35.3\,$\pm$\,3.2 & \textbf{81.8\,$\pm$\,1.7} & 71.5\,$\pm$\,4.2 & 78.1\,$\pm$\,3.4 & 50.0\,$\pm$\,0.5 & \textbf{49.9\,$\pm$\,1.1} & 50.5\,$\pm$\,0.5 \\
\rowcolor{fsshade}\fedstop{} (geometric) & \textbf{63.5\,$\pm$\,0.7} & \textbf{39.0\,$\pm$\,2.3} & 81.1\,$\pm$\,1.9 & 73.1\,$\pm$\,5.2 & 77.0\,$\pm$\,3.6 & \textbf{50.1\,$\pm$\,0.4} & 49.5\,$\pm$\,1.4 & \textbf{50.8\,$\pm$\,0.7} \\
\bottomrule
\end{tabular}
\end{table}

\section{Discussion}
\label{sec:discussion}

\textbf{Limitations.} Three limitations matter most. First, optimality requires stopping decisions that do not reverse, and local progress is more strongly autocorrelated than the margin of Corollary~\ref{cor:correlated} allows. An exact trajectory-aware rule gains nothing over the one-stage rule (Table~\ref{tab:scorecard}), but the guarantee does not cover these trajectories. Second, the model describes a single round, so it predicts that budgets anneal but not how fast they do. Third, the drift price is motivated by an aggregation bound but chosen on validation data, because the curvature constants that would fix it are unavailable for neural networks. Beyond the theory, the evidence comes from small vision and language models in simulated federations of 100 clients, so larger models and cross-device scale remain untested.

\textbf{Future directions.} Three directions follow. One is optimal stopping under dependent progress: rules with regret guarantees for autocorrelated gains, between one-stage look-ahead and full dynamic programming. The other is a cross-round theory that couples each client's stopping problem to the server's trajectory, so that the annealing of local work, and with it the drift price, is derived rather than tuned. The third is stopping without monitoring passes, by correcting the losses recorded during training for within-epoch drift, which would remove most of the rule's remaining overhead.

\section{Conclusion}

We recast the choice of local epochs in federated learning as optimal stopping with a priced drift cost. This view yields a provably optimal one-stage threshold, closed forms that make the rule's behaviour interpretable, and an exact account of what estimation costs. Tested against epoch-level evidence, the model captures how local training actually behaves, and its central lesson is that the stage of training, not the identity of the client, determines how much local work pays. Acting on this, \fedstop{} protects the worst-off clients at a cost of at most one point of mean accuracy, using a decision rule whose computation and memory are negligible beside training. It also degrades gracefully, rather than failing, when its single parameter is set too high.

\section*{Broader Impact Statement}

\fedstop{} aims to make federated learning serve its worst-off clients better: under strong data heterogeneity it raises the accuracy of the least-served decile of clients at a cost of at most one point of mean accuracy. This can benefit users whose data differ most from the majority, who are often the ones a shared model serves poorly. The rule has three costs that deployments should weigh. First, where budgets vary, clients train for more local epochs than under tuned FedAvg, which consumes more energy and battery on client devices. Monitoring on a small fixed subset of examples reduces this overhead. Second, the gain for the worst-off clients can come with a slight loss for the best-served ones. Third, as in FedNova, each client reports its number of local steps, which reveals its local dataset size and how long it trained. \fedstop{} exchanges no data or gradients beyond these standard updates, so it adds no other privacy exposure. All experiments use public benchmark datasets and involve no human subjects.

\bibliography{fedstop}

@inproceedings{mcmahan2017, author={McMahan, H. B. and Moore, E. and Ramage, D. and Hampson, S. and Ag{\"u}era y Arcas, B.}, title={Communication-efficient learning of deep networks from decentralized data}, booktitle={Proceedings of the 20th International Conference on Artificial Intelligence and Statistics (AISTATS)}, pages={1273--1282}, year={2017}}

@inproceedings{li2020fedprox, author={Li, T. and Sahu, A. K. and Zaheer, M. and Sanjabi, M. and Talwalkar, A. and Smith, V.}, title={Federated optimization in heterogeneous networks}, booktitle={Proceedings of Machine Learning and Systems (MLSys)}, year={2020}}

@inproceedings{karimireddy2020, author={Karimireddy, S. P. and Kale, S. and Mohri, M. and Reddi, S. and Stich, S. and Suresh, A. T.}, title={{SCAFFOLD}: Stochastic controlled averaging for federated learning}, booktitle={Proceedings of the 37th International Conference on Machine Learning (ICML)}, pages={5132--5143}, year={2020}}

@inproceedings{wang2020fednova, author={Wang, J. and Liu, Q. and Liang, H. and Joshi, G. and Poor, H. V.}, title={Tackling the objective inconsistency problem in heterogeneous federated optimization}, booktitle={Advances in Neural Information Processing Systems (NeurIPS)}, year={2020}}

@article{wang2019adaptive, author={Wang, S. and Tuor, T. and Salonidis, T. and Leung, K. K. and Makaya, C. and He, T. and Chan, K.}, title={Adaptive federated learning in resource constrained edge computing systems}, journal={IEEE Journal on Selected Areas in Communications}, volume={37}, number={6}, pages={1205--1221}, year={2019}}

@inproceedings{stich2019, author={Stich, S. U.}, title={Local {SGD} converges fast and communicates little}, booktitle={International Conference on Learning Representations (ICLR)}, year={2019}}

@inproceedings{khaled2020, author={Khaled, A. and Mishchenko, K. and Richt{\'a}rik, P.}, title={Tighter theory for local {SGD} on identical and heterogeneous data}, booktitle={Proceedings of the 23rd International Conference on Artificial Intelligence and Statistics (AISTATS)}, pages={4519--4529}, year={2020}}

@inproceedings{woodworth2020, author={Woodworth, B. and Patel, K. K. and Srebro, N.}, title={Minibatch vs local {SGD} for heterogeneous distributed learning}, booktitle={Advances in Neural Information Processing Systems (NeurIPS)}, year={2020}}

@article{qin2023, author={Qin, T. and Etesami, S. R. and Uribe, C. A.}, title={The role of local steps in local {SGD}}, journal={Optimization Methods and Software}, year={2023}}

@inproceedings{charles2021surrogate, author={Charles, Z. and Kone{\v{c}}n{\'y}, J.}, title={Convergence and accuracy trade-offs in federated learning and meta-learning}, booktitle={Proceedings of the 24th International Conference on Artificial Intelligence and Statistics (AISTATS)}, pages={2575--2583}, year={2021}}

@inproceedings{patel2024limits, author={Patel, K. K. and Glasgow, M. and Zindari, A. and Wang, L. and Stich, S. U. and Cheng, Z. and Joshi, N. and Srebro, N.}, title={The limits and potentials of local {SGD} for distributed heterogeneous learning with intermittent communication}, booktitle={Proceedings of the 37th Conference on Learning Theory (COLT)}, pages={4115--4157}, year={2024}}

@inproceedings{shenaj2025alt, author={Shenaj, D. and Belilovsky, E. and Zanuttigh, P.}, title={Adaptive local training in federated learning}, booktitle={Proceedings of the 33rd European Signal Processing Conference (EUSIPCO)}, pages={1907--1911}, year={2025}}

@inproceedings{panchal2023flash, author={Panchal, K. and Choudhary, S. and Mitra, S. and Mukherjee, K. and Sarkhel, S. and Mitra, S. and Guan, H.}, title={Flash: Concept drift adaptation in federated learning}, booktitle={Proceedings of the 40th International Conference on Machine Learning (ICML)}, pages={26931--26962}, year={2023}}

@inproceedings{jiang2020optimal, author={Jiang, P. and Ying, L.}, title={An optimal stopping approach for iterative training in federated learning}, booktitle={Proceedings of the 54th Annual Conference on Information Sciences and Systems (CISS)}, year={2020}}

@article{mahmoudi2022fedcau, author={Mahmoudi, A. and Ghadikolaei, H. S. and {Da Silva J{\'u}nior}, J. M. B. and Fischione, C.}, title={{FedCau}: A proactive stop policy for communication and computation efficient federated learning}, journal={IEEE Transactions on Wireless Communications}, year={2024}}

@article{niu2024flrce, author={Niu, Z. and Dong, H. and Qin, A. K. and Gu, T.}, title={{FLrce}: Resource-efficient federated learning with early-stopping strategy}, journal={IEEE Transactions on Mobile Computing}, volume={23}, number={12}, pages={14514--14529}, year={2024}}

@inproceedings{li2020qffl, author={Li, T. and Sanjabi, M. and Beirami, A. and Smith, V.}, title={Fair resource allocation in federated learning}, booktitle={International Conference on Learning Representations (ICLR)}, year={2020}}

@inproceedings{mohri2019afl, author={Mohri, M. and Sivek, G. and Suresh, A. T.}, title={Agnostic federated learning}, booktitle={Proceedings of the 36th International Conference on Machine Learning (ICML)}, year={2019}}

@book{chow1971, author={Chow, Y. S. and Robbins, H. and Siegmund, D.}, title={Great Expectations: The Theory of Optimal Stopping}, publisher={Houghton Mifflin}, address={Boston, MA}, year={1971}}

@book{peskir2006, author={Peskir, G. and Shiryaev, A.}, title={Optimal Stopping and Free-Boundary Problems}, publisher={Birkh{\"a}user}, address={Basel}, year={2006}}

@inproceedings{karimi2016pl, author={Karimi, H. and Nutini, J. and Schmidt, M.}, title={Linear convergence of gradient and proximal-gradient methods under the {Polyak-{\L}ojasiewicz} condition}, booktitle={Joint European Conference on Machine Learning and Knowledge Discovery in Databases (ECML PKDD)}, pages={795--811}, year={2016}}

@article{liu2022pl, author={Liu, C. and Zhu, L. and Belkin, M.}, title={Loss landscapes and optimization in over-parameterized non-linear systems and neural networks}, journal={Applied and Computational Harmonic Analysis}, volume={59}, pages={85--116}, year={2022}}

@article{bottou2018optimization, author={Bottou, L. and Curtis, F. E. and Nocedal, J.}, title={Optimization methods for large-scale machine learning}, journal={SIAM Review}, volume={60}, number={2}, pages={223--311}, year={2018}}

@article{kaplan2020scaling, author={Kaplan, J. and McCandlish, S. and Henighan, T. and Brown, T. B. and Chess, B. and Child, R. and Gray, S. and Radford, A. and Wu, J. and Amodei, D.}, title={Scaling laws for neural language models}, journal={arXiv preprint arXiv:2001.08361}, year={2020}}

@inproceedings{li2022niidbench, author={Li, Q. and Diao, Y. and Chen, Q. and He, B.}, title={Federated learning on non-{IID} data silos: An experimental study}, booktitle={Proceedings of the 38th IEEE International Conference on Data Engineering (ICDE)}, year={2022}}

@article{caldas2018leaf, author={Caldas, S. and Duddu, S. M. K. and Wu, P. and Li, T. and Kone{\v{c}}n{\'y}, J. and McMahan, H. B. and Smith, V. and Talwalkar, A.}, title={{LEAF}: A benchmark for federated settings}, journal={arXiv preprint arXiv:1812.01097}, year={2018}}
\bibliographystyle{tmlr}

\clearpage
\appendix
\setcounter{table}{0}\setcounter{figure}{0}
\renewcommand{\thetable}{S\arabic{table}}\renewcommand{\thefigure}{S\arabic{figure}}
\renewcommand{\theHtable}{S\arabic{table}}\renewcommand{\theHfigure}{S\arabic{figure}}

\section*{Supplementary material}
\suppressfloats[t]
These appendices report every configuration of the prespecified study that the main text does not tabulate: SCAFFOLD, further baseline configurations, ablations of the power-law rule, budget controls and variants on the naturally partitioned data. It also reports the details of the client-side stopping rules ALT and FLASH, training dynamics and the allocation of local epochs over clients and rounds, the measured cost of loss monitoring with a fixed example subset, the synthetic benchmark that preceded the real-data comparison, the predictability of client workload, the shape estimator's settings, and diagnostics computed from the epoch logs. All results use the main text's protocol.

\section{Protocol}
Every number is computed from the stored records of the reporting seeds 6--10, paired with geometric or power-law \fedstop{} by partition, initialisation and client sampling. Intervals are two-sided 95\% Student $t$ intervals on four degrees of freedom, unadjusted for multiple comparisons. Tail is the label-reweighted worst decile on CIFAR-10 and the cross-fitted worst decile on FEMNIST and Shakespeare. Collapsed runs, those ending at chance accuracy, are retained and scored at chance, and ``Coll.'' counts them. Paired contrasts are omitted (--) for configurations with collapsed runs, where they are uninformative. Unless stated, configurations share the main text's learning rate, rounds, batch size, $\Emax=20$ and step-normalised aggregation.

\section{SCAFFOLD}
SCAFFOLD \citep{karimireddy2020} corrects the direction of local updates with control variates. \fedstop{} chooses how long local training lasts, so the two address different questions and could be combined. It was run once per setting at the shared learning rate, without the hyperparameter selection that every other tuned method received (Table~\ref{tab:s-scaffold}). It collapsed on all five CIFAR-10 and four of five FEMNIST reporting seeds, a known fragility of control variates when clients participate rarely (here about every tenth round) and train for many steps. On Shakespeare it completed every seed and is the most accurate method tested, $4.77$ points above geometric \fedstop{} ($[4.67,4.87]$) with a tail $2.74$ points higher ($[0.53,4.95]$). A tuned comparison, selecting its server step size on seeds 1--5, would show whether the collapses are intrinsic.

\begin{table}[t]
\caption{SCAFFOLD, untuned, at the shared learning rate (\%, five seeds)}
\label{tab:s-scaffold}
\centering\footnotesize
\setlength{\tabcolsep}{2.2pt}
\begin{tabular}{@{}lcccl@{}}
\toprule
Setting & Accuracy & Tail & Collapsed & Per-seed accuracy \\
\midrule
CIFAR-10, $\alpha=0.1$ & 10.0\,$\pm$\,0.0 & 10.0\,$\pm$\,0.0 & 5/5 & 10.0, 10.0, 10.0, 10.0, 10.0 \\
FEMNIST & 17.6\,$\pm$\,35.7 & 15.0\,$\pm$\,30.0 & 4/5 & 1.6, 1.6, 1.6, 1.6, 81.4 \\
Shakespeare & 54.9\,$\pm$\,0.5 & 52.2\,$\pm$\,1.3 & 0/5 & 55.1, 54.6, 55.4, 54.3, 55.3 \\
\bottomrule
\end{tabular}
\end{table}

\section{Additional baseline configurations}
Table~\ref{tab:s-baselines} lists every FedNova budget, every configuration of the adaptive interval of \citet{wang2019adaptive}, and FedProx \citep{li2020fedprox} on CIFAR-10. FedNova \citep{wang2020fednova} matches the tail of geometric \fedstop{} at $E=5$, $10$ and $20$, at two to eight times its epochs and lower mean accuracy. Wang et al.'s algorithm appears in two variants. Option A, the one in the main text, adapts the aggregation interval in epochs with the cap $\Emax$, matching the epoch-budget question the paper studies. Option C follows the original resource model more closely, trading local computation against aggregation with a cost ratio $b/c=10.11$ and a larger interval cap (100 instead of $\Emax$). With its original $\phi$, option C is level with \fedstop{} on CIFAR-10. With the $\phi$ of option A it trains longer for a lower mean. FedProx's proximal weight was selected from five values on seeds 1--5, and one reporting seed collapsed.

\begin{table}[t]
\caption{Baseline configurations on CIFAR-10, $\alpha=0.1$ ($\Delta$: configuration minus geometric \fedstop{})}
\label{tab:s-baselines}
\centering\scriptsize
\setlength{\tabcolsep}{2.2pt}
\begin{tabular}{@{}lcccccc@{}}
\toprule
Configuration & Accuracy & Tail & Epochs ($10^3$) & Coll. & $\Delta$ accuracy & $\Delta$ tail \\
\midrule
FedNova, $E=1$ & 61.4\,$\pm$\,2.3 & 30.4\,$\pm$\,6.3 & 10.0 & 0 & $-2.17$ [$-4.35$, $+0.01$] & $-8.56$ [$-14.81$, $-2.31$] \\
FedNova, $E^\star=2$ & 63.1\,$\pm$\,2.2 & 34.8\,$\pm$\,3.4 & 20.0 & 0 & $-0.46$ [$-2.44$, $+1.51$] & $-4.22$ [$-8.29$, $-0.15$] \\
FedNova, $E=5$ & 62.7\,$\pm$\,0.6 & 38.9\,$\pm$\,2.8 & 50.0 & 0 & $-0.89$ [$-1.52$, $-0.26$] & $-0.03$ [$-2.05$, $+2.00$] \\
FedNova, $E=10$ & 60.9\,$\pm$\,0.9 & 39.0\,$\pm$\,3.0 & 100.0 & 0 & $-2.61$ [$-3.47$, $-1.75$] & $-0.02$ [$-4.62$, $+4.57$] \\
FedNova, $E=20$ & 59.2\,$\pm$\,1.1 & 38.9\,$\pm$\,3.3 & 200.0 & 0 & $-4.33$ [$-5.38$, $-3.28$] & $-0.07$ [$-5.72$, $+5.58$] \\
Wang et al.\ (A) & 62.0\,$\pm$\,1.1 & 32.7\,$\pm$\,3.0 & 20.5 & 0 & $-1.58$ [$-2.58$, $-0.57$] & $-6.29$ [$-8.18$, $-4.39$] \\
Wang et al.\ (A), step-normalised & 63.1\,$\pm$\,2.1 & 35.1\,$\pm$\,2.9 & 20.7 & 0 & $-0.43$ [$-2.43$, $+1.58$] & $-3.83$ [$-8.60$, $+0.95$] \\
Wang et al.\ (C), original $\phi$ & 64.5\,$\pm$\,2.1 & 39.9\,$\pm$\,8.0 & 36.5 & 0 & $+1.00$ [$-2.23$, $+4.23$] & $+0.90$ [$-8.00$, $+9.79$] \\
Wang et al.\ (C), $\phi$ of A & 61.0\,$\pm$\,2.9 & 38.5\,$\pm$\,6.8 & 72.4 & 0 & $-2.52$ [$-6.53$, $+1.50$] & $-0.49$ [$-10.99$, $+10.00$] \\
FedProx & 51.6\,$\pm$\,23.3 & 26.8\,$\pm$\,9.5 & 20.0 & 1 & -- & -- \\
\bottomrule
\end{tabular}
\end{table}

\section{Ablations of the power-law rule}
Each variant changes one setting of the power-law rule, the prespecified primary, and is compared with it in Table~\ref{tab:s-ablations}:
\begin{itemize}
\item \emph{AR(1) rule}: continuation value of the AR(1) cutoff policy, solved by backward induction on a 201-point residual grid with 20-node Gauss--Hermite noise, using the shrunk scale. Against the one-stage rule with the same shrunk scale, which isolates the change of rule, it differs by $+0.07$ ($[-0.57,0.72]$) in accuracy and $+0.15$ ($[-2.00,2.29]$) in tail.
\item \emph{Sequential probability ratio test}: a CUSUM-form test on the net gain ($\alpha=\beta=0.05$).
\item \emph{Upper-confidence-bound rule}: continue while the upper confidence bound of the windowed mean net gain is positive (window 3, $\alpha=0.05$).
\item \emph{Linear multi-step look-ahead}: maximise the cumulative predicted net gain over all later epochs, with a drift term linear in the observed squared drift.
\item \emph{Minimum of two epochs}: $k_{\min}=2$.
\item \emph{Shrunk loss scale}: the loss scale is shrunk towards a causal prediction from earlier rounds.
\item \emph{Window} $w=1$ and \emph{all participations}: the shape is pooled over one or every past participation instead of four.
\item \emph{Escalating} and \emph{de-escalating drift price}: $\lambda$ follows a geometric schedule across rounds spanning a factor of four around the selected value.
\item \emph{Plain} and \emph{epoch-normalised aggregation}: unnormalised averaging, or normalisation by epochs rather than steps.
\end{itemize}
No variant improves on the power-law rule's tail with an interval excluding zero. The sequential and upper-confidence-bound rules trained far longer and collapsed on five and four of five seeds. The exact AR(1) rule matches the one-stage rule.

\begin{table}[t]
\caption{Prespecified ablations of the power-law rule on CIFAR-10, $\alpha=0.1$ ($\Delta$: variant minus the power-law rule)}
\label{tab:s-ablations}
\centering\scriptsize
\setlength{\tabcolsep}{2.2pt}
\begin{tabular}{@{}lcccccc@{}}
\toprule
Configuration & Accuracy & Tail & Epochs ($10^3$) & Coll. & $\Delta$ accuracy & $\Delta$ tail \\
\midrule
AR(1) rule & 62.5\,$\pm$\,2.1 & 33.5\,$\pm$\,2.3 & 13.3 & 0 & $-0.23$ [$-1.01$, $+0.56$] & $-0.28$ [$-3.61$, $+3.05$] \\
Sequential probability ratio test & 10.0\,$\pm$\,0.0 & 6.0\,$\pm$\,5.5 & 130.2 & 5 & -- & -- \\
Upper-confidence-bound rule & 10.4\,$\pm$\,0.8 & 8.0\,$\pm$\,4.5 & 48.9 & 4 & -- & -- \\
Linear multi-step look-ahead & 61.2\,$\pm$\,2.6 & 29.9\,$\pm$\,6.8 & 10.0 & 0 & $-1.58$ [$-3.71$, $+0.55$] & $-3.89$ [$-7.67$, $-0.10$] \\
Minimum of two epochs & 63.3\,$\pm$\,1.2 & 35.6\,$\pm$\,2.4 & 20.1 & 0 & $+0.53$ [$-0.16$, $+1.21$] & $+1.83$ [$-1.37$, $+5.04$] \\
Shrunk loss scale & 62.5\,$\pm$\,1.8 & 33.4\,$\pm$\,3.6 & 13.3 & 0 & $-0.30$ [$-0.85$, $+0.24$] & $-0.42$ [$-2.34$, $+1.49$] \\
Window $w=1$ & 63.2\,$\pm$\,1.1 & 33.5\,$\pm$\,5.1 & 14.3 & 0 & $+0.40$ [$-0.89$, $+1.69$] & $-0.30$ [$-1.22$, $+0.63$] \\
Window over all participations & 61.9\,$\pm$\,2.2 & 31.6\,$\pm$\,4.8 & 11.5 & 0 & $-0.85$ [$-2.12$, $+0.41$] & $-2.24$ [$-4.23$, $-0.25$] \\
Escalating drift price & 62.9\,$\pm$\,2.1 & 34.7\,$\pm$\,4.1 & 14.0 & 0 & $+0.08$ [$-1.05$, $+1.21$] & $+0.90$ [$-0.83$, $+2.63$] \\
De-escalating drift price & 61.3\,$\pm$\,1.8 & 30.5\,$\pm$\,5.3 & 14.6 & 0 & $-1.44$ [$-2.68$, $-0.19$] & $-3.32$ [$-6.35$, $-0.29$] \\
Plain aggregation & 61.2\,$\pm$\,0.9 & 30.0\,$\pm$\,4.6 & 13.6 & 0 & $-1.57$ [$-3.55$, $+0.41$] & $-3.78$ [$-8.71$, $+1.16$] \\
Epoch-normalised aggregation & 61.5\,$\pm$\,1.8 & 30.2\,$\pm$\,6.0 & 13.4 & 0 & $-1.32$ [$-3.31$, $+0.66$] & $-3.62$ [$-11.32$, $+4.08$] \\
\bottomrule
\end{tabular}
\end{table}

\section{Budget controls}
Table~\ref{tab:s-controls} replays the power-law rule's budgets under four rearrangements and reports two non-deployable references. \emph{Random swap within round} hands each round's budgets to its clients in random order. \emph{Random swap within size strata} does so only among clients of similar size (five strata ranked by steps per epoch). \emph{Permuted across rounds} permutes each client's budgets across its own participations, removing the trend over training. \emph{De-annealed} fixes each client's budget at its mean over the run, matching total epochs exactly. The \emph{hindsight oracle} is the study's hindsight-utility reference, and the \emph{round-wise validation oracle} chooses each round's budget by global validation loss. Decomposing the de-annealed contrast, the trend over rounds contributes $+1.61 [-3.68, +6.91]$ tail points and the per-client allocation of a constant budget $+2.30 [-3.21, +7.81]$, against $+3.91 [+2.58, +5.25]$ in total.

The main text's equal split (Table~\ref{tab:swap}) replays geometric \fedstop{}'s per-round totals on CIFAR-10 and FEMNIST and divides each round's epochs equally among its clients. Paired epoch totals are identical, but because clients differ in size the equal split executes $4.8\%$ more SGD steps on CIFAR-10 ($[3.5,6.1]$) and $1.5\%$ more on FEMNIST ($[1.1,1.9]$), and trains on $5.1\%$ and $1.8\%$ more examples. Geometric \fedstop{} has the higher tail on one of five seeds in each setting, and on the shared-index FEMNIST tail the two differ by $-1.13$ ($[-3.60,1.34]$). As in the main text, these intervals establish neither superiority nor equivalence.

\begin{table}[t]
\caption{Controls built from power-law budgets on CIFAR-10, $\alpha=0.1$ ($\Delta$: control minus the power-law rule)}
\label{tab:s-controls}
\centering\scriptsize
\setlength{\tabcolsep}{2.2pt}
\begin{tabular}{@{}lcccccc@{}}
\toprule
Configuration & Accuracy & Tail & Epochs ($10^3$) & Coll. & $\Delta$ accuracy & $\Delta$ tail \\
\midrule
Random swap within round & 63.9\,$\pm$\,1.5 & 36.3\,$\pm$\,2.7 & 13.3 & 0 & $+1.11$ [$-0.03$, $+2.24$] & $+2.54$ [$-0.11$, $+5.19$] \\
Random swap within size strata & 62.8\,$\pm$\,1.4 & 32.7\,$\pm$\,5.6 & 13.3 & 0 & $-0.02$ [$-1.04$, $+1.00$] & $-1.08$ [$-2.72$, $+0.57$] \\
Permuted across rounds & 63.3\,$\pm$\,0.9 & 32.2\,$\pm$\,8.5 & 13.3 & 0 & $+0.52$ [$-2.09$, $+3.14$] & $-1.61$ [$-6.91$, $+3.68$] \\
De-annealed & 61.8\,$\pm$\,2.4 & 29.9\,$\pm$\,4.7 & 13.3 & 0 & $-0.99$ [$-2.43$, $+0.44$] & $-3.91$ [$-5.25$, $-2.58$] \\
Hindsight oracle (ref.) & 60.6\,$\pm$\,2.3 & 29.8\,$\pm$\,6.0 & 200.0 & 0 & $-2.21$ [$-3.93$, $-0.49$] & $-3.96$ [$-6.23$, $-1.69$] \\
Round-wise validation oracle (ref.) & 64.0\,$\pm$\,0.5 & 37.8\,$\pm$\,6.5 & 200.0 & 0 & $+1.22$ [$-0.84$, $+3.27$] & $+4.00$ [$-4.10$, $+12.09$] \\
\bottomrule
\end{tabular}
\end{table}

\section{Training dynamics}
The main text's training-curve figure follows top-1 and worst-decile accuracy over all $1000$ rounds for geometric \fedstop{}, tuned FedAvg, a random swap of \fedstop{}'s budgets and per-client early stopping. On FEMNIST the swap learns slightly faster in the first 200 rounds, consistent with its larger share of SGD steps, and the methods converge to similar top-1 accuracy after about round 400, while tuned FedAvg's worst decile stays about four points below \fedstop{}'s. Per-client early stopping ends at chance accuracy on every CIFAR-10 seed. On two seeds it never learns, and on three it falls back after peaking at $27$--$43\%$, because returning its best iterate lets a client whose validation loss worsens hand back the unchanged broadcast model.

\begin{figure}[t]
\centering
\includegraphics[width=\textwidth]{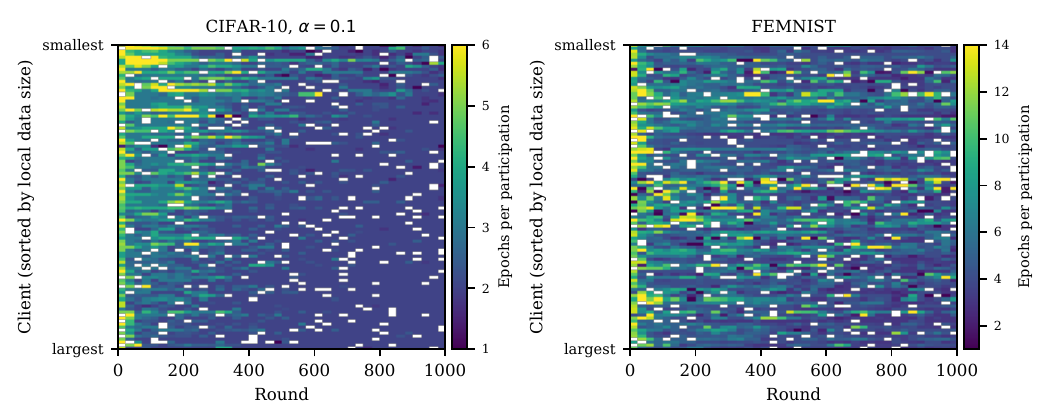}
\caption{Epochs per participation chosen by geometric \fedstop{} for every client and 25-round window (reporting seed 6), with clients sorted by local data size. White cells are windows in which a client was not sampled.}
\label{fig:s-budgets}
\end{figure}

Figure~\ref{fig:s-budgets} shows where geometric \fedstop{} spends its epochs, for every client and round of one reporting seed, with clients sorted by local data size. On CIFAR-10 budgets fall from left to right for almost every client, so the stage of training sets the budget (P3), and the smallest clients train somewhat longer early on, the weak size effect of P4. On FEMNIST budgets also start high everywhere but later vary more from round to round within each client, consistent with the round explaining less of the budget variance there.

\section{Variants on the naturally partitioned data}
Table~\ref{tab:s-leaf} compares two variants with geometric \fedstop{} on FEMNIST and Shakespeare. Forcing a second epoch costs accuracy on Shakespeare, where the rule chooses one epoch. A drift price transferred from CIFAR-10 at $\alpha=0.5$ stays level with the locally selected one.

\begin{table}[t]
\caption{Variants on FEMNIST and Shakespeare ($\Delta$: variant minus geometric \fedstop{})}
\label{tab:s-leaf}
\centering\scriptsize
\setlength{\tabcolsep}{2.2pt}
\resizebox{\linewidth}{!}{%
\begin{tabular}{@{}lcccccc@{}}
\toprule
Configuration & Accuracy & Tail & Epochs ($10^3$) & Coll. & $\Delta$ accuracy & $\Delta$ tail \\
\midrule
FEMNIST: Minimum of two epochs & 81.6\,$\pm$\,1.6 & 72.4\,$\pm$\,5.1 & 35.0 & 0 & $+0.50$ [$-1.38$, $+2.37$] & $-0.71$ [$-4.00$, $+2.58$] \\
FEMNIST: Drift price transferred from CIFAR-10, $\alpha=0.5$ & 81.8\,$\pm$\,1.7 & 71.2\,$\pm$\,4.9 & 10.3 & 0 & $+0.78$ [$-1.46$, $+3.01$] & $-1.97$ [$-6.23$, $+2.29$] \\
Shakespeare: Minimum of two epochs & 48.4\,$\pm$\,0.4 & 47.5\,$\pm$\,1.3 & 10.6 & 0 & $-1.76$ [$-1.89$, $-1.64$] & $-2.04$ [$-3.14$, $-0.94$] \\
Shakespeare: Drift price transferred from CIFAR-10, $\alpha=0.5$ & 50.1\,$\pm$\,0.5 & 50.4\,$\pm$\,1.9 & 5.0 & 0 & $-0.07$ [$-0.19$, $+0.05$] & $+0.86$ [$-1.69$, $+3.40$] \\
\bottomrule
\end{tabular}}
\end{table}

\section{Shape estimator and pilot priors}
The shape minimises $J(\theta)=n_{\mathrm{res}}\log(\mathrm{SSE}(\theta)/n_{\mathrm{res}})+n_{\mathrm{p}}(\theta-\theta_0)^2/s_0^2$ over the current and last $w=4$ participations, on 200 grid points refined by golden section, with $n_{\mathrm{p}}=2$ and search ranges $\rho\in[0.05,0.9995]$, $a\in[0.1,4]$. The prior $(\theta_0,s_0)$ and the drift prior $\delta^{\mathrm{prior}}$, the pilot median of $\delta(1)$, come from a pilot round per setting and profile. The prior is used alone until three informative rounds exist. Table~\ref{tab:s-priors} lists the values with the selected drift price and the share of decisions whose shape equals its prior. Each drift price is the best of a five-value grid spanning the range that the pilot round calibrates, selected on seeds 1--5 by validation accuracy. The selected prices range from $6.7\times10^{-4}$ to $1.27\times10^{4}$. Each grid spans many orders of magnitude, because the budget depends on the drift price only through $\log\kappa$. On CIFAR-10, for instance, the geometric grid runs from $1.27\times10^{4}$ down to $1.3\times10^{-23}$. The power-law exponent rarely leaves its prior, because a rule that seldom trains past one epoch seldom accumulates informative rounds.

\begin{table}[t]
\caption{Shape-estimator settings and pilot priors}
\label{tab:s-priors}
\centering\footnotesize
\setlength{\tabcolsep}{2.2pt}
\begin{tabular}{@{}lccccc@{}}
\toprule
Setting, profile & $\theta_0$ & $s_0$ & $\delta^{\mathrm{prior}}$ & $\lambda$ & Shape at prior \\
\midrule
CIFAR-10 ($\alpha=0.1$), geometric & 0.302 & 0.160 & 0.118 & 0.029 & 5.3\% \\
CIFAR-10 ($\alpha=0.1$), power-law & 2.426 & 0.561 & 0.118 & 1.39 & 69.1\% \\
CIFAR-10 ($\alpha=0.5$), geometric & 0.302 & 0.160 & 0.118 & $1.27\!\times\!10^{4}$ & -- \\
CIFAR-10 ($\alpha=0.5$), power-law & 2.426 & 0.561 & 0.118 & 1.39 & -- \\
CIFAR-100 ($\alpha=0.1$), geometric & 0.221 & 0.116 & 1.52 & 798 & -- \\
CIFAR-100 ($\alpha=0.1$), power-law & 2.742 & 0.616 & 1.52 & 1.34 & -- \\
FEMNIST, geometric & 0.482 & 0.285 & 0.208 & $6.7\!\times\!10^{-4}$ & 5.8\%$^{*}$ \\
FEMNIST, power-law & 1.924 & 0.787 & 0.208 & 0.151 & -- \\
Shakespeare, geometric & 0.296 & 0.074 & 398 & 0.104 & -- \\
Shakespeare, power-law & 2.482 & 0.292 & 398 & $1.5\!\times\!10^{-3}$ & 57.6\% \\
\bottomrule
\multicolumn{6}{@{}l}{\scriptsize $^{*}$One reporting seed. --: not computed, as it requires epoch-level logs.}\\
\end{tabular}
\end{table}

\section{Client-side stopping rules}
Table~\ref{tab:s-ruleparams} lists the thresholds selected for ALT and FLASH in the main text's comparison. ALT's grid starts at the 10th, 30th, 60th and 90th percentiles of the per-batch feature cosine similarity in a five-epoch FedAvg pilot, and it is extended in the coordinate $\log((1+\tau)/(1-\tau))$. FLASH's grid starts at 0.1, 0.3, 1 and 3 times the pilot's median first-epoch drop in validation loss, and it is extended by factors of three. FLASH uses the mean validation cross-entropy, including the evaluation before the first epoch. Selection uses seeds 1--5, the last five validation evaluations and a calibrated-work tie-break. Diverged selection runs are scored at chance, and seven of ALT's CIFAR-10 selection runs diverged. The pilot calibrates the grids only.

Table~\ref{tab:s-rules} adds the secondary results, with every difference taken as geometric \fedstop{} minus the rule. It gives the Holm-adjusted $p$-values over the six primary tail contrasts, the fixed-cohort tails, each rule's epochs per participation and the modeled work over the full horizon. Geometric \fedstop{} executes $2.49$, $5.37$ and $1.00$ epochs per participation on the three settings, with modeled work of $31.4$, $61.4$ and $5.72$ thousand client-epoch equivalents. It therefore uses more modeled work than both rules on CIFAR-10 and Shakespeare. On FEMNIST it uses less than ALT, whose criterion rarely fires, and more than FLASH, which stops after about one epoch.

\begin{table}[t]
\caption{Selected thresholds of the client-side rules}
\label{tab:s-ruleparams}
\centering\footnotesize
\begin{tabular}{@{}lcccccc@{}}
\toprule
& \multicolumn{3}{c}{ALT} & \multicolumn{3}{c}{FLASH} \\
\cmidrule(lr){2-4}\cmidrule(lr){5-7}
Setting & $\tau$ & Ext. & At edge & $\gamma$ & Ext. & At edge \\
\midrule
CIFAR-10 & 0.9889 & 2 & yes & 0.160 & 1 & no \\
FEMNIST & 0.8928 & 0 & no & 0.645 & 1 & no \\
Shakespeare & 0.9947 & 2 & yes & 0.331 & 2 & yes \\
\bottomrule
\end{tabular}
\end{table}

\begin{table}[t]
\caption{Client-side rules: secondary results ($\Delta$: geometric \fedstop{} minus the rule, 95\% CI, unadjusted, with the rule's epochs per participation and work)}
\label{tab:s-rules}
\centering\scriptsize
\setlength{\tabcolsep}{3pt}
\begin{tabular}{@{}llcccccc@{}}
\toprule
Setting & Rule & $\Delta$ top-1 & $\Delta$ worst decile & Holm $p$ & $\Delta$ fixed cohort & Epochs & Work ($10^3$) \\
\midrule
CIFAR-10 & ALT & $+2.16$ [$-0.06$, $4.38$] & $+8.95$ [$1.40$, $16.50$] & 0.151 & -- & 1.19 & 14.1 \\
CIFAR-10 & FLASH & $+0.38$ [$-0.62$, $1.38$] & $+3.68$ [$1.52$, $5.83$] & 0.055 & -- & 2.06 & 21.3 \\
FEMNIST & ALT & $-0.50$ [$-2.89$, $1.88$] & $-0.03$ [$-4.39$, $4.34$] & 1.000 & $-1.73$ [$-4.92$, $1.45$] & 19.07 & 214.0 \\
FEMNIST & FLASH & $-0.73$ [$-2.95$, $1.49$] & $+1.68$ [$-1.84$, $5.19$] & 1.000 & $-1.11$ [$-3.15$, $0.93$] & 1.06 & 10.9 \\
Shakespeare & ALT & $+0.15$ [$0.01$, $0.30$] & $-0.16$ [$-1.85$, $1.54$] & 1.000 & $+0.40$ [$0.07$, $0.74$] & 1.00 & 5.36 \\
Shakespeare & FLASH & $+0.11$ [$-0.15$, $0.37$] & $-0.36$ [$-2.07$, $1.35$] & 1.000 & $+0.28$ [$-0.35$, $0.91$] & 1.01 & 5.32 \\
\bottomrule
\end{tabular}
\end{table}

\section{Computational cost and subset monitoring}
Compute is counted in modeled client-epoch equivalents, including every monitoring pass and weighting a forward pass by each setting's forward-to-training cost ratio. It is not measured elapsed time. In the reported runs each client evaluates its training loss on its full local dataset before training and after every epoch. Table~\ref{tab:s-subset} compares this with monitoring on a fixed subset of $\min(128,n_i)$ local examples, measured on the reporting seeds with every selected drift price unchanged. Work is expressed as a percentage of two-epoch FedAvg's work on the same seeds, both to reach $95\%$ of tuned FedAvg's final mean accuracy and over the full horizon.

\begin{figure}[t]
\centering
\includegraphics[width=\textwidth]{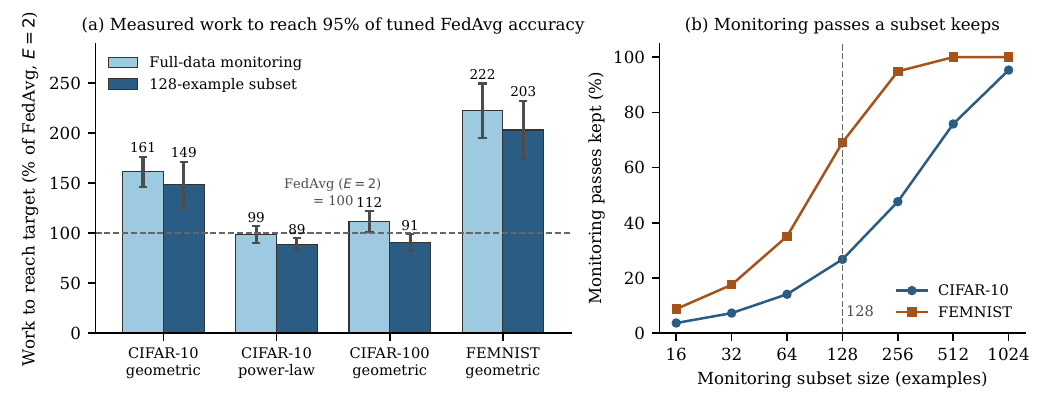}
\caption{Subset monitoring. (a) Measured work to reach $95\%$ of tuned FedAvg's final accuracy, as a percentage of two-epoch FedAvg's work (mean $\pm$ one standard deviation over five seeds). (b) Share of monitoring passes kept by a fixed subset of $\min(s,n_i)$ examples, computed from the client sizes and epoch counts of the reporting runs.}
\label{fig:s-cost}
\end{figure}

The subset lowers the work in every setting. It brings the power-law rule on CIFAR-10 ($89\%$) and the geometric rule on CIFAR-100 ($91\%$) below two-epoch FedAvg's work to reach the target, while geometric \fedstop{} on CIFAR-10 and FEMNIST still needs more ($149\%$ and $203\%$). The saving is smaller than an accounting estimate that holds the decisions fixed would suggest, partly because the subset changes some decisions: on CIFAR-10, geometric clients executed $88$ more epochs per run ($[11,165]$). No mean-accuracy difference excludes zero. The geometric CIFAR-10 tail fell by $1.34$ points ($[0.10,2.57]$), which triggered a prespecified re-selection of the drift price. The re-selection chose the same price, so its repeat on the same seeds is a same-policy rerun rather than a tuning gain, and it is not an independent replication. In that repeat the tail differs from full monitoring by $-0.20$ ($[-2.51,2.12]$). Figure~\ref{fig:s-cost} shows the measured work and, for reference, the share of monitoring passes that a subset of each size keeps.

\begin{table}[t]
\caption{Monitoring on a fixed 128-example subset (work in \% of two-epoch FedAvg, $\Delta$: subset minus full monitoring, 95\% CI)}
\label{tab:s-subset}
\centering\scriptsize
\setlength{\tabcolsep}{2.5pt}
\resizebox{\linewidth}{!}{%
\begin{tabular}{@{}lcccccccc@{}}
\toprule
& \multicolumn{3}{c}{Work to reach target} & \multicolumn{3}{c}{Work over full horizon} & & \\
\cmidrule(lr){2-4}\cmidrule(lr){5-7}
Setting, profile & Full & Subset & $\Delta$ (points) & Full & Subset & $\Delta$ (points) & $\Delta$ top-1 & $\Delta$ worst decile \\
\midrule
CIFAR-10, geometric & 161.3 & 148.6 & $-12.8$ [$-25.3$, $-0.3$] & 157.1 & 140.8 & $-16.4$ [$-17.1$, $-15.6$] & $-0.31$ [$-0.88$, $0.27$] & $-1.34$ [$-2.57$, $-0.10$] \\
CIFAR-10, power-law & 98.5 & 89.2 & $-9.4$ [$-14.4$, $-4.3$] & 88.3 & 77.3 & $-10.9$ [$-11.4$, $-10.5$] & $-0.37$ [$-1.62$, $0.88$] & $-1.22$ [$-3.43$, $0.99$] \\
CIFAR-100, geometric & 111.6 & 91.0 & $-20.7$ [$-23.2$, $-18.2$] & 67.7 & 55.2 & $-12.5$ [$-12.6$, $-12.4$] & $+0.15$ [$-0.20$, $0.50$] & $+0.24$ [$-0.00$, $0.47$] \\
FEMNIST, geometric & 222.4 & 203.3 & $-19.0$ [$-57.1$, $19.1$] & 307.2 & 298.2 & $-9.0$ [$-12.4$, $-5.6$] & $-0.42$ [$-0.89$, $0.04$] & $-0.56$ [$-4.44$, $3.31$] \\
CIFAR-10, geometric, same-$\lambda$ repeat & 161.3 & 138.1 & $-23.2$ [$-35.8$, $-10.6$] & 157.1 & 140.9 & $-16.3$ [$-17.0$, $-15.5$] & $+0.02$ [$-0.94$, $0.98$] & $-0.20$ [$-2.51$, $2.12$] \\
\bottomrule
\end{tabular}}
\end{table}

\section{Synthetic benchmark}
Before the real-data comparison, we compared the rules on a synthetic benchmark. A 32--64--10 MLP is trained on a ten-class Gaussian mixture with three sub-clusters per class, over 100 clients with a Dirichlet label split ($\alpha=0.1$), for 120 rounds at $\eta=0.05$. FLASH's client rule, ALT's released criterion and an equal split of \fedstop{}'s budgets are each tuned on two seeds and reported on five (Table~\ref{tab:s-synth}). Compute counts a forward pass as one third of an epoch. \fedstop{} improves on one-epoch FedAvg by $3.4$ points ($[0.9,5.9]$) and by $5.2$ points in the worst decile ($[2.6,7.8]$), but it is statistically level with every other method. FLASH's rule is nominally the most accurate and uses less than half of \fedstop{}'s compute, much of which is loss monitoring. ALT's similarity signal rarely fires in this small network, whose representations stay above $0.98$ cosine similarity, so it trains about 18 epochs per client and is the most expensive. The equal split matches \fedstop{} ($-0.9$ points, $[-3.2,1.3]$) with $30\%$ less compute, because it does not monitor.

\begin{table}[t]
\caption{Synthetic benchmark (\%, five seeds, compute in $10^3$ client-epoch equivalents)}
\label{tab:s-synth}
\centering\footnotesize
\setlength{\tabcolsep}{3pt}
\begin{tabular}{@{}lcccc@{}}
\toprule
Method & Accuracy & Worst decile & Epochs & Compute \\
\midrule
FedAvg, $E=1$ & 57.0\,$\pm$\,1.4 & 34.6\,$\pm$\,4.5 & 1.0 & 1.2 \\
FLASH client rule & 61.5\,$\pm$\,2.3 & 41.4\,$\pm$\,9.4 & 2.3 & 2.9 \\
ALT & 59.4\,$\pm$\,1.1 & 41.1\,$\pm$\,3.7 & 18.2 & 29.1 \\
\fedstop{} (geometric) & 60.4\,$\pm$\,1.2 & 39.7\,$\pm$\,4.4 & 3.8 & 6.4 \\
Equal split (control) & 61.4\,$\pm$\,1.2 & 40.8\,$\pm$\,6.5 & 3.8 & 4.5 \\
\bottomrule
\end{tabular}
\end{table}

\section{Predicting client workload}
A client's budget times its known steps per epoch is its local workload, so a server that can predict budgets can anticipate computation and stragglers. Predicting each client's budget at its next participation from its previous one gives a mean absolute error of $0.30$ epochs on CIFAR-10 and $1.44$ on FEMNIST, against $0.47$ and $2.16$ for the previous round's cohort mean, and is exact for $76\%$ and $48\%$ of participations. Budget history is thus a useful scheduling signal on CIFAR-10 and a modest one on FEMNIST, consistent with budgets varying more across rounds than across clients.

\section{Diagnostics from the epoch logs}
Table~\ref{tab:s-diag} collects the quantities behind the main text's discussion of the plug-in and the model's limits. The deployed drift estimate includes the first epoch, so the first decision prices continuation well above the drift that later epochs realise. On CIFAR-10 the geometric shape sits at its lower search bound in most decisions. The margin of Corollary~5 holds in no client-round with at least four epochs, and under the fitted autoregressive model a stopping decision would reverse in about a fifth to a quarter of them. These in-sample estimates on short trajectories are indicative only. On FEMNIST, successive loss decreases decline faster than the drift ratio $\sigma\approx0.83$ in most epochs, as the geometric declining-drift corollary requires. In seven of the ten setting--profile pairs the selected drift price is the largest on its grid: the power-law rule on CIFAR-10 at $\alpha=0.1$, and both rules on CIFAR-10 at $\alpha=0.5$, CIFAR-100 and Shakespeare. A large price makes extra epochs expensive, which is why \fedstop{} chooses one epoch in most of these settings. Only geometric \fedstop{} on CIFAR-10 at $\alpha=0.1$ and both rules on FEMNIST select an interior price.

\begin{table}[t]
\caption{Diagnostics computed from the epoch logs of the reporting runs}
\label{tab:s-diag}
\centering\footnotesize
\setlength{\tabcolsep}{2.2pt}
\begin{tabular}{@{}p{0.50\linewidth}cc@{}}
\toprule
Quantity & CIFAR-10 & FEMNIST \\
\midrule
First-decision drift price / later realised drift rate (median) & 16.2 & 6.9 \\
Client-rounds stopped after one epoch & 2.1\% & 5.1\% \\
Fitted $\hat\rho$: median [IQR] & 0.05 [0.05, 0.12] & 0.18 [0.08, 0.36] \\
$\hat\rho$ at a search bound & 65.1\% & 15.1\% \\
Pooled lag-one correlation $\phi$ of progress surprises & $-0.40$ & $-0.32$ \\
Client-rounds ($\ge4$ epochs) meeting the margin of Corollary~5 & 0.0\% & 0.0\% \\
Decisions that would reverse under the fitted model & 18.7\% & 26.9\% \\
Successive-decrease ratio, epochs 2--4: median and share $\le0.83$ & -- & 0.34 and 86.3\% \\
Median local training examples per client & 390 & 144 \\
Monitoring work kept by a fixed $\min(128,n_i)$ subset & 26.8\% & 69.1\% \\
\bottomrule
\end{tabular}
\end{table}

\end{document}